\documentclass[11pt]{article}

\usepackage[preprint]{acl}

 \usepackage{booktabs}
\usepackage{times}
\usepackage{latexsym}
\usepackage{amsmath}
\usepackage[T1]{fontenc}
\usepackage[utf8]{inputenc}
\usepackage{float}
\usepackage{microtype}
\usepackage{inconsolata}
\usepackage{caption}
\usepackage{placeins}
\usepackage{multirow}
\usepackage{booktabs}
\usepackage{graphicx}
\usepackage{multirow}
\usepackage{tikz}
\usepackage{amssymb}
\usetikzlibrary{arrows.meta, positioning, calc}
\definecolor{cBlue}{RGB}{55,110,196}
\definecolor{cTeal}{RGB}{0,140,125}
\definecolor{cOrange}{RGB}{215,130,15}
\definecolor{cPurple}{RGB}{120,50,160}
\definecolor{cRed}{RGB}{190,55,55}
\definecolor{cGray}{RGB}{110,110,110}
\definecolor{cDark}{RGB}{40,40,40}
\newcommand*\circled[1]{\tikz[baseline=(char.base)]{
            \node[shape=circle,draw,inner sep=2pt] (char) {#1};}}

\definecolor{gold}{RGB}{248, 187, 87}  
\definecolor{silver}{RGB}{158, 158, 158}  
\definecolor{bronze}{RGB}{206, 115, 50}  
\usepackage{fontawesome5}

\title{AILS-NTUA at SemEval-2026 Task 8: Evaluating Multi-Turn RAG Conversations}

\title{AILS-NTUA at SemEval-2026 Task 8: Evaluating Multi-Turn RAG Conversations}

\author{
\textbf{Dimosthenis Athanasiou}, \textbf{Maria Lymperaiou}, \textbf{Giorgos Filandrianos},\\
\textbf{Athanasios Voulodimos}, \textbf{Giorgos Stamou}\\
National Technical University of Athens\\
\href{mailto:dimosthenis.athanas@gmail.com}{\texttt{dimosthenis.athanas@gmail.com}},
\href{mailto:marialymp@ails.ece.ntua.gr}{\texttt{\{marialymp,}}
\href{mailto:geofila@ails.ece.ntua.gr}{\texttt{geofila\}@ails.ece.ntua.gr}},\\
\href{mailto:thanosv@mail.ntua.gr}{\texttt{thanosv@mail.ntua.gr}},
\href{mailto:gstam@cs.ntua.gr}{\texttt{gstam@cs.ntua.gr}}
}

\begin{document}
\maketitle
\begin{abstract}
We present the AILS-NTUA system for SemEval-2026 Task~8 (MTRAGEval),
addressing all three subtasks of multi-turn retrieval-augmented generation:
passage retrieval~(A), reference-grounded response generation~(B),
and end-to-end RAG~(C).
Our unified architecture is built on two principles:
(i)~a \textit{query-diversity-over-retriever-diversity} strategy,
where five complementary LLM-based query reformulations are issued to a
single corpus-aligned sparse retriever and fused via variance-aware
nested Reciprocal Rank Fusion; and
(ii)~a multistage generation pipeline that decomposes grounded generation
into evidence span extraction, dual-candidate drafting, and
calibrated multi-judge selection.
Our system ranks \textbf{1st in Task~A} (nDCG@5:~0.5776,
$+$20.5\% over the strongest baseline) and \textbf{2nd in Task~B}
(HM:~0.7698).
Empirical analysis shows that query diversity over a well-aligned
retriever outperforms heterogeneous retriever ensembling,
and that answerability calibration---rather than retrieval
coverage—is the primary bottleneck in end-to-end performance.
\end{abstract}

\section{Introduction}
Large Language Models (LLMs) have revolutionized several 
aspects of NLP, from human-like language generation~\cite{tian-etal-2024-large-language} to complex reasoning~\cite{giadikiaroglou-etal-2024-puzzle} and scientific 
tasks \cite{kao-etal-2024-solving, song2025evaluatinglargelanguagemodels, han-etal-2025-generalist}.
However, the static nature of such knowledge does not allow for updates without the re-training burden, rendering a state-of-the-art LLM outdated within a few years. At the same time, LLMs are knowledge-rich but opaque systems that cannot attribute their knowledge of specific facts to a well-defined source, raising concerns regarding their trustworthiness.

Retrieval Augmented Generation (RAG) addresses both 
limitations by dynamically consulting external sources 
at inference time \cite{rag}. By conditioning generation 
on retrieved documents, RAG enhances factual accuracy, 
robustness to knowledge cut-offs, and accountability 
of claims, becoming an established paradigm for knowledge-intensive tasks \cite{izacard-grave-2021-leveraging, menick2022teachinglanguagemodelssupport, gao2024retrievalaugmentedgenerationlargelanguage}.

In practice, RAG systems have demonstrated significant gains in dense and hybrid retrieval, multi-document fusion and  grounded generation through citations and self-verification \cite{arivazhagan-etal-2023-hybrid, asai2024selfrag}. Yet, their success relies on independent user queries, without interfering with past information within a conversation. In real-life scenarios though, conversations unfold within multi-turn interactions, suggesting the treatment of user queries in the context of the dialog and previously retrieved information. Consequently, RAG systems must become context-aware to avoid out-of-context augmentation and error accumulation across turns.

In light of such demands, \textit{multi-turn RAG} is attracting increasing attention, empowering context-aware question reformulation and dialog-conditioned generation. Prior work has explored multi-turn conversational reasoning \cite{zheng2023judging}, active retrieval \cite{jiang-etal-2023-active}, unanswerability \cite{faithdial} and long-form conversational RAG \cite{ikat, kuo-etal-2025-rad}, motivating the development of the  MTRAG benchmark \cite{mt-rag}. Unlike other RAG benchmarks, MTRAG features diverse question types, together with answerable, unanswerable, partial, and conversational questions and relevant and irrelevant passages, spanning four domains in a multi-turn setup.

In this paper, we approach multi-turn RAG as a retrieval stability problem: across turns, small reformulation errors accumulate, progressively degrading evidence grounding and leading to confident hallucinations. We design an architecture that explicitly separates context preservation, evidence discovery, and answer commitment. Rather than relying on retriever ensembles or larger models, we stabilize retrieval through diversity-controlled rewriting over a single corpus-aligned index. We stabilize generation by deferring answer commitment until agreement is reached across multiple evidence judges. In summary, our contributions are:
\circled{1} A stability-oriented multi-strategy rewriting method that improves recall without degrading top-k precision by controlling reformulation variance through nested Reciprocal Rank 
Fusion (RRF) aggregation.
\circled{2} An agentic evidence-commitment generation pipeline that reduces conversational hallucinations by separating extraction, drafting, and selection into agreement-driven stages.
\circled{3} A unified analysis of multi-turn retrieval failure modes on MTRAG, showing that:
   (a) retrieval diversity helps depth recall but harms early precision,
   (b) cross-encoders complement but cannot replace aligned sparse retrieval,
   (c) LLM-based reranking saturates once grounding quality is sufficient.
Our system ranks 1st on Task~A and 2nd on Task~B, validating the proposed design principles.
\section{Background}
\paragraph{Task description} MTRAG comprises 110 conversations, each with user--agent
history $H_t = \{(u_i, a_i)\}_{i=1}^{t-1}$ (where $u_i$,
$a_i$ are the user utterance and agent response at turn $i$,
and $q_t \equiv u_t$ the current query) and an associated
document corpus. Evaluation is decomposed into three subtasks. \circled{1} \textbf{Task~A - Retrieval Only} provides the user/agent conversation, along with the corresponding document corpus. The output should contain the 10 most relevant passages from the corpus, ordered according to similarity to the last query. \circled{2} \textbf{Task~B - Generation with Reference Passages (Reference)} again provides the user/agent conversation, but contrary to Task~A the relevant query passages are provided. The goal is to generate a passage-grounded agent response to the last user query. \circled{3} \textbf{Task~C: Generation with Retrieved Passages (RAG)} receives the same inputs as Task~A. Then, participants have to retrieve up to 10 relevant passages and use them to generate an appropriate response to the last user query.
A thorough data exploration is provided in App.~\ref{sec:eda}.

\paragraph{Related work} RAG has served as a widely adopted framework for conditioning generation on external knowledge, enabling parameter-free knowledge updates \cite{rag}. Subsequent work explored multi-passage architectures for integrating multiple relevant contexts \cite{izacard-grave-2021-leveraging}, alongside retriever–-generator training for enhanced retrieval awareness \cite{atlas}, retrieval augmentation for black-box models \cite{shi-etal-2024-replug}, and self-reflective mechanisms to improve grounding \cite{asai2024selfrag}. In a parallel avenue, conversational research has formalized multi-turn retrieval under context-dependent queries \cite{dalton2020treccast2019conversational}, often via question rewriting into standalone queries \cite{rewrite, open-domain-question-answering-goes-conversational}, with later work addressing topic switching and context drift \cite{adlakha-etal-2022-topiocqa}. The integration of retrieval and generation for multi-turn interactions has led to the introduction of self-checking strategies tailored to conversational QA \cite{convrag} and active retrieval methods  for adapting evidence selection during generation or across turns \cite{jiang-etal-2023-active}. At the same time, unanswerability emerges as a critical challenge for faithful information-seeking dialogue \cite{faithdial}, while evaluation targets long-form, retrieval-augmented dialogues \cite{ikat, kuo-etal-2025-rad}. Finally, automated evaluation frameworks for RAG have been proposed to diagnose retrieval quality and grounding-related failure modes \cite{es-etal-2024-ragas}.

\section{System Overview}
\begin{figure*}[t!]
\centering
\resizebox{0.9\textwidth}{!}{%
\begin{tikzpicture}[
    >={Stealth[length=1.6mm, width=1mm]},
    stage/.style={
        rectangle, rounded corners=2pt, draw=#1!60, line width=0.5pt,
        fill=#1!8, inner sep=2pt, minimum width=2.8cm,
        font=\sffamily\fontsize{6.5}{8}\selectfont\bfseries, text=cDark, align=center
    },
    smallbox/.style={
        rectangle, rounded corners=1.5pt, draw=#1!35, line width=0.35pt,
        fill=white, inner sep=2pt,
        font=\sffamily\fontsize{5.5}{7}\selectfont, text=cDark
    },
    iobox/.style={
        rectangle, rounded corners=2pt, draw=#1!65, line width=0.7pt,
        fill=#1!10, inner sep=2pt,
        font=\sffamily\fontsize{6.5}{8}\selectfont\bfseries, text=cDark
    },
    lbl/.style={font=\sffamily\fontsize{5}{6}\selectfont, text=cGray},
    arr/.style={->, line width=0.45pt, #1!55},
    darr/.style={->, line width=0.35pt, #1!40, dashed},
    ptitle/.style={font=\sffamily\fontsize{7.5}{9}\selectfont\bfseries, text=#1},
]

\begin{scope}[xshift=0cm]

\node[ptitle=cBlue] (tA) at (1.5,0) {(a) Task A: Retrieval};

\node[iobox=cBlue, below=0.15cm of tA] (Ain) {Query + History};

\node[stage=cOrange, below=0.18cm of Ain] (As1) {1.\ Multi-Strategy Rewriting};

\node[smallbox=cOrange, below=0.13cm of As1, xshift=-0.85cm] (r1) {Minimal};
\node[smallbox=cOrange, right=0.06cm of r1] (r2) {Corpus-Sp.};
\node[smallbox=cOrange, right=0.06cm of r2] (r3) {HyDE};
\node[smallbox=cOrange, below=0.04cm of r1, xshift=0.42cm] (r4) {CoT};
\node[smallbox=cOrange, right=0.06cm of r4] (r5) {Anchor-KW};

\node[stage=cTeal, below=0.18cm of r4, xshift=0.1cm] (As2) {2.\ Sparse Retrieval};
\node[lbl, below=0.02cm of As2] (As2lbl) {$5\!\times\!$top-100 per query};

\node[stage=cBlue, below=0.18cm of As2lbl] (As3) {3.\ Hybrid Reranking};
\node[smallbox=cBlue, below=0.12cm of As3, xshift=-0.45cm] (h1) {Retriever};
\node[smallbox=cBlue, right=0.08cm of h1] (h2) {Cross-Enc.};
\node[lbl, right=0.06cm of h2] {w-RRF};

\node[stage=cPurple, below=0.18cm of h1, xshift=0.45cm] (As4) {4.\ Nested RRF Fusion};
\node[lbl, below=0.02cm of As4] (As4lbl) {corpus-specific weights};

\node[iobox=cTeal, below=0.18cm of As4lbl] (Aout) {Top-\textit{k} Passages};

\draw[arr=cBlue] (Ain) -- (As1);
\draw[arr=cOrange] (As1) -- (r2);
\draw[arr=cOrange] (r4.south) -- ++(0,-0.06) -| ([xshift=-0.15cm]As2.north);
\draw[arr=cOrange] (r5.south) -- ++(0,-0.06) -| ([xshift=0.15cm]As2.north);
\draw[arr=cTeal] (As2lbl.south) -- ++(0,0.01) -- (As3.north);
\draw[arr=cBlue] (As3) -- (h1);
\draw[arr=cBlue] (As3) -- (h2);
\draw[arr=cBlue] (h1.south) -- ++(0,-0.06) -| ([xshift=-0.15cm]As4.north);
\draw[arr=cBlue] (h2.south) -- ++(0,-0.06) -| ([xshift=0.15cm]As4.north);
\draw[arr=cPurple] (As4lbl.south) -- ++(0,0.01) -- (Aout.north);
\end{scope}

\begin{scope}[xshift=4.8cm]

\node[ptitle=cTeal] (tB) at (1.5,0) {(b) Task B: Generation};

\node[iobox=cTeal, below=0.15cm of tB] (Bin) {Reference Passages};

\node[stage=cRed, below=0.18cm of Bin] (Bs0) {1.\ Answerability Check};
\node[smallbox=cRed, right=0.12cm of Bs0] (idk) {\textit{refusal}};
\draw[darr=cRed] (Bs0) -- (idk);

\node[stage=cBlue, below=0.18cm of Bs0] (Bs1) {2.\ Evidence Span Extraction};
\node[lbl, below=0.02cm of Bs1] (Bs1lbl) {up to $M$ verbatim sentences};

\node[stage=cPurple, below=0.18cm of Bs1lbl] (Bs2) {3.\ Dual Candidate Generation};
\node[smallbox=cPurple, below=0.12cm of Bs2, xshift=-0.45cm] (v1) {Cand.\ A};
\node[smallbox=cPurple, right=0.1cm of v1] (v2) {Cand.\ B};

\node[stage=cOrange, below=0.18cm of v1, xshift=0.45cm] (Bs3) {4.\ Judge-Based Selection};
\node[smallbox=cOrange, below=0.12cm of Bs3, xshift=-0.48cm] (j1) {Technical};
\node[smallbox=cOrange, right=0.08cm of j1] (j2) {User Sat.};
\node[lbl, below=0.04cm of j1, xshift=0.48cm] (jlbl) {+ extractiveness shaping};

\node[stage=cGray, below=0.18cm of jlbl] (Bs4) {5.\ Micro-Adjustments};

\node[iobox=cTeal, below=0.18cm of Bs4] (Bout) {Final Response};

\draw[arr=cTeal] (Bin) -- (Bs0);
\draw[arr=cRed] (Bs0) -- (Bs1);
\draw[arr=cBlue] (Bs1lbl.south) -- ++(0,0.01) -- (Bs2.north);
\draw[arr=cPurple] (Bs2) -- (v1);
\draw[arr=cPurple] (Bs2) -- (v2);
\draw[arr=cPurple] (v1.south) -- ++(0,-0.06) -| ([xshift=-0.15cm]Bs3.north);
\draw[arr=cPurple] (v2.south) -- ++(0,-0.06) -| ([xshift=0.15cm]Bs3.north);
\draw[arr=cOrange] (Bs3) -- (j1);
\draw[arr=cOrange] (Bs3) -- (j2);
\draw[arr=cOrange] (j1.south) -- ++(0,-0.07) -| ([xshift=-0.12cm]jlbl.north);
\draw[arr=cOrange] (j2.south) -- ++(0,-0.07) -| ([xshift=0.12cm]jlbl.north);
\draw[arr=cGray] (jlbl.south) -- ++(0,0.01) -- (Bs4.north);
\draw[arr=cGray] (Bs4) -- (Bout);
\end{scope}

\begin{scope}[xshift=9.6cm]

\node[ptitle=cPurple] (tC) at (1.5,0) {(c) Task C: End-to-End RAG};

\node[iobox=cBlue, below=0.15cm of tC] (Cin) {Query + History};

\node[stage=cBlue, below=0.18cm of Cin] (Cs1) {Task A Pipeline (\S3.1)};
\node[lbl, below=0.02cm of Cs1] (Cs1lbl) {retrieve top-$N$ passages};

\node[stage=cPurple, below=0.18cm of Cs1lbl] (Cs2) {Span Extr.\ + Generation};

\node[stage=cRed, below=0.18cm of Cs2] (Cs3) {Multi-Judge Answerability};
\node[smallbox=cRed, below=0.12cm of Cs3, xshift=-0.85cm] (cj1) {Document};
\node[smallbox=cRed, right=0.06cm of cj1] (cj2) {Span};
\node[smallbox=cRed, right=0.06cm of cj2] (cj3) {Answer};

\node[stage=cPurple, below=0.18cm of cj2] (Cs4) {Confidence-Weighted Vote};

\node[smallbox=cTeal, below=0.15cm of Cs4, xshift=-0.85cm] (cans) {Answerable};
\node[smallbox=cRed, right=0.35cm of cans] (cunan) {Unanswerable};

\node[iobox=cTeal, below=0.15cm of cans] (CgenB) {Task B (\S3.2)};
\node[iobox=cRed, below=0.15cm of cunan] (Cref) {Refusal};

\node[iobox=cPurple, below=0.2cm of CgenB, xshift=0.85cm] (Cout) {Final Response};

\draw[arr=cBlue] (Cin) -- (Cs1);
\draw[arr=cBlue] (Cs1lbl.south) -- ++(0,0.01) -- (Cs2.north);
\draw[arr=cPurple] (Cs2) -- (Cs3);
\draw[arr=cRed] (Cs3) -- (cj1);
\draw[arr=cRed] (Cs3) -- (cj2);
\draw[arr=cRed] (Cs3) -- (cj3);
\draw[arr=cRed] (cj1.south) -- ++(0,-0.06) -| ([xshift=-0.25cm]Cs4.north);
\draw[arr=cRed] (cj2) -- (Cs4);
\draw[arr=cRed] (cj3.south) -- ++(0,-0.06) -| ([xshift=0.25cm]Cs4.north);
\draw[arr=cPurple] (Cs4) -- (cans);
\draw[arr=cPurple] (Cs4) -- (cunan);
\draw[arr=cTeal] (cans) -- (CgenB);
\draw[arr=cRed] (cunan) -- (Cref);
\draw[arr=cTeal] (CgenB.south) -- ++(0,-0.07) -| ([xshift=-0.12cm]Cout.north);
\draw[arr=cRed] (Cref.south) -- ++(0,-0.07) -| ([xshift=0.12cm]Cout.north);
\end{scope}

\end{tikzpicture}
}
\caption{System architecture: (a)~Task~A retrieval pipeline, 
(b)~Task~B generation pipeline, (c)~Task~C end-to-end RAG 
with answerability gate.}
\label{fig:system}
\end{figure*}
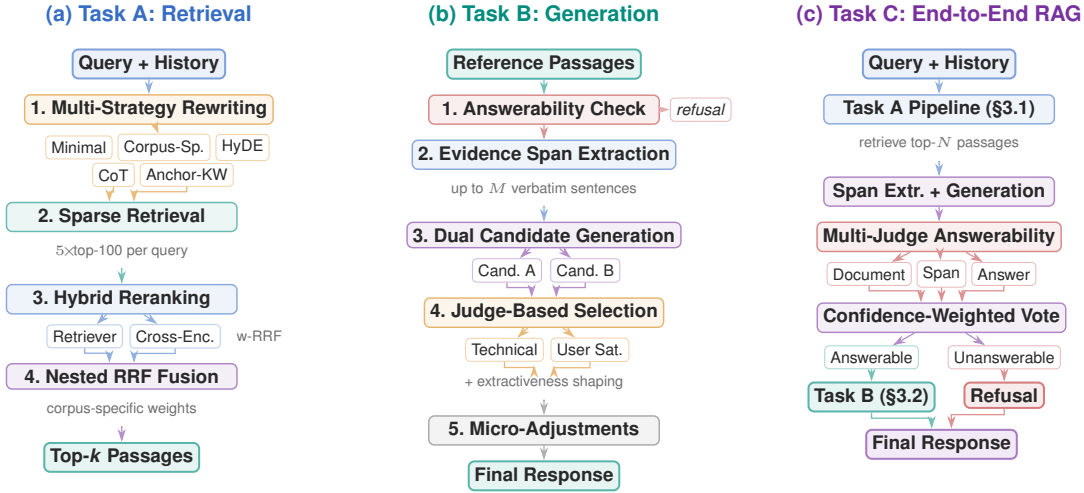

We address all three subtasks with a unified architecture. For retrieval (Task~A), we treat conversational search primarily as a query formulation problem and aggregate diverse reformulations over a single corpus-aligned index. For generation (Tasks~B and~C), we adopt an agentic approach that decomposes grounded response generation into answerability detection, evidence identification, candidate drafting, and evidence-aware selection.

\paragraph{Task~A: Multi-Strategy Retrieval} Our retrieval system addresses multi-turn conversational search
through a four-stage pipeline (Figure~\ref{fig:system}a).

\noindent\circled{1}~\textbf{Query Rewriting.}
Given conversation history $H = \{(u_1, a_1), \ldots,
(u_{t-1}, a_{t-1})\}$ and the current user query $q_t$, we generate
five complementary reformulations, each targeting distinct
failure modes: \textit{Minimal}---resolves coreferences and conversational omissions;
\textit{Corpus-Specific}---adapts the query to domain terminology;   
\textit{Hypothetical Document Embedding}---generates a hypothetical answer
passage, bridging the query--document vocabulary gap~\cite{gao2023hyde}; \textit{Chain-of-Thought (CoT)}---expands the information need through step-wise reasoning; \textit{Anchor-Keyword}---extracts salient
entities and keywords optimized for sparse lexical matching.
All strategies use XML-formatted prompts with instructions and few-shot examples (App.~\ref{app:taska_prompts}).
\circled{2}~\textbf{Retrieval.}
Each reformulation is issued to the retriever, producing five ranked lists
$\mathcal{R}_1, \ldots, \mathcal{R}_5$.
\circled{3}~\textbf{Hybrid Reranking.}
We fuse 
retriever and reranker rankings via weighted RRF:
\begin{equation}
\mathrm{Score}(d) = \frac{1}{k + r_\mathrm{E}(d)} +
\alpha \cdot \frac{1}{k + r_\mathrm{R}(d)}
\label{eq:hybrid_rrf}
\end{equation}
where $r_\mathrm{E}$ and $r_\mathrm{R}$ denote retriever 
and reranker ranks, $k$ controls rank decay, and $\alpha$ 
balances the two signals.
\circled{4}~\textbf{Nested Fusion.}
\textit{Minimal} and \textit{Corpus-Specific} produce stable, 
high-precision rankings, while the remaining three exhibit higher variance 
but complementary coverage. We propose a two-level fusion: 
\textbf{Level~1} pre-aggregates the three high-variance 
strategies into a \textit{Weak Consensus} ranking; 
\textbf{Level~2} combines this with the two stable strategies 
via corpus-specific weighted RRF. The final score for each 
passage $d$ is:
\begin{equation}
\mathrm{Score}_{\textit{final}}(d) = \!\sum_{s} w^{(c)}_s \cdot
\frac{1}{k^{(c)} + \mathrm{rank}_s(d)}
\label{eq:nested_rrf}
\end{equation}
where $s$ ranges over the three Level~2 inputs and 
$w^{(c)}_s$, $k^{(c)}$ are corpus-dependent parameters 
(Table~\ref{tab:nested_rrf_params}). The top-$k$ passages 
by $\mathrm{Score}_{\textit{final}}$ form the retrieval output.

\paragraph{Task~B: Agentic Generation Pipeline} We treat grounded response generation as a multi-stage 
decision process (Figure~\ref{fig:system}b).

\noindent\circled{1}~\textbf{Answerability Classification.}
Each turn is classified based on context as \emph{unanswerable} (empty context$\rightarrow$ triggering a 
calibrated refusal) or \emph{answerable/partial} (proceeding 
to generation).
\circled{2}~\textbf{Evidence Span Extraction.}
An extraction module identifies verbatim supporting sentences 
from provided passages. These spans replace full passages as generator input.
\circled{3}~\textbf{Dual Candidate Generation.}
Two response candidates are generated, conditioned on extracted
spans, conversation history, and
question-type-specific length guidance, with strict grounding 
constraints.
\circled{4}~\textbf{Judge-Based Selection.}
A \textit{Technical Judge} evaluates faithfulness and completeness against extracted spans, while a \textit{User Satisfaction Judge}, invoked on a stochastic subset of turns, evaluates naturalness. Selection combines judge preferences with an \textit{Extractiveness Shaping} term that penalizes unsupported abstraction and verbatim copying, and discourages refusal when evidence exists.
\circled{5}~\textbf{Micro-Adjustments.}
A light post-editing pass addresses residual length and
extractiveness bound violations, and removes formulaic hedging
phrases (details in
App.~\ref{app:selection_score}).

\paragraph{Task~C: End-to-End RAG} Task~C integrates retrieval and generation while additionally deciding
whether retrieved evidence justifies answering. Unlike Tasks~A and~B,
retrieved passages may be irrelevant and answerability is often ambiguous.

Our system retrieves the top-5 passages via the Task~A pipeline, extracts spans, and generates two candidate
answers. Three specialized LLM judges evaluate answerability from complementary views:
a \textbf{document judge} checks passage relevance,
a \textbf{span judge} verifies evidence coverage,
and an \textbf{answer judge} evaluates response adequacy.

An arbiter aggregates the binary verdicts using confidence-weighted voting
with dissenter override. Answerable turns continue through the Task~B
pipeline; otherwise a calibrated refusal is produced.

\section{Experimental setup}
\paragraph{Dataset}
We evaluate on the MTRAG benchmark~\cite{mt-rag}:
110 multi-turn conversations (842 turns) across four domains
(ClapNQ, FiQA, Govt, Cloud).
Documents are segmented into 512-token passages (stride 100) and
indexed using Elasticsearch Learned Sparse Encoder (ELSER~v1). Task~A uses the 777 answerable and partially answerable tasks; Tasks~B and~C use all 842 of the development set, on which tuning is conducted; results are reported on held-out test set\cite{Rosenthal2026MTRAGEval,
rosenthal2026mtragunbenchmarkopenchallenges}.
\begin{table}[h]
\centering
\small
\resizebox{\columnwidth}{!}{%
\begin{tabular}{lcccc}
\toprule
\textbf{Characteristic} & \textbf{ClapNQ} & \textbf{FiQA} & \textbf{Govt} & \textbf{Cloud} \\
\midrule
Documents & 4,293 & 57,638$^\dagger$ & 7,661 & 8,578 \\
Passages & 183,408 & 61,022 & 49,607 & 72,442 \\
Avg.\ passages/doc & 42.7 & 1.1 & 6.5 & 8.4 \\
Domain & Wikipedia & Finance & Government & Tech docs \\
\bottomrule
\end{tabular}
}
\caption{MTRAG corpus statistics. $^\dagger$FiQA contains 
individual forum posts.}
\label{tab:corpus_stats}
\end{table}


\paragraph{Task~A Configuration.}
Rewriting strategies use DeepSeek-V3.2 ($\tau{=}0.0$) with 
6 user and 3 assistant turns as context. Each 
reformulation retrieves top-100 passages from ELSER~v1, 
selected after benchmarking 9 alternatives (App.~\ref{app:retriever_comparison}). Hybrid reranking 
uses Cohere Rerank~v4 with $k{=}60$ and $\alpha{=}0.5$ 
(Eq.~\ref{eq:hybrid_rrf}). The weak-consensus group aggregates HyDE, CoT, and Anchor-Keyword with equal weights ($k_{\text{internal}}{=}40$); final fusion uses corpus-specific weights tuned on dev set (Table~\ref{tab:nested_rrf_params}); uniform weights yield less than 1\% R@5 degradation, 
confirming robustness to weight selection (App.~\ref{app:nested_rrf}).

\begin{table}[]
\centering
\small
\begin{tabular}{lcccc}
\toprule
\textbf{Parameter} & \textbf{ClapNQ} & \textbf{FiQA} & \textbf{Govt} & \textbf{Cloud} \\
\midrule
$k_{\text{final}}$ & 20 & 60 & 40 & 20 \\
$w_{\text{Minimal}}$ & 0.55 & 0.45 & 0.65 & 0.65 \\
$w_{\text{Corpus-Spec}}$ & 0.40 & 0.40 & 0.25 & 0.30 \\
$w_{\text{WeakCons}}$ & 0.05 & 0.15 & 0.10 & 0.05 \\
\bottomrule
\end{tabular}
\caption{\textit{Nested RRF} parameters per domain.}

\label{tab:nested_rrf_params}
\end{table}

\paragraph{Task~B Configuration.}
We route pipeline stages to three models: DeepSeek-V3.2 for span extraction ($\tau{=}0.0$, up to 8 spans) and technical judging; GPT-4o for dual candidate generation ($\tau_1{=}0.0$, $\tau_2{=}0.1$); and GPT-4o-mini for the \textit{User Satisfaction Judge} (60\% invocation rate) and micro-adjustments ($\tau{=}0.1$). Target length is question-type dependent (App.~\ref{app:qtype}). Selection combines technical scores, user preference, and
extractiveness calibration (App.~\ref{app:selection_score}).

\paragraph{Task~C Configuration.}
All three judges and the arbiter use GPT-4o ($\tau{=}0.0$). Refusal responses are capped
at 25~words. Answerable turns reuse the full Task~B pipeline
configuration. 
\paragraph{Evaluation Metrics}
For Task~A, we report Recall@$k$ and nDCG@$k$ 
($k \in \{5, 10\}$). For Tasks~B and~C, we use 
three metrics from the MTRAG evaluation 
framework~\cite{mt-rag}: $\text{RB}_{\text{alg}}$ 
(reference-based algorithmic), 
$\text{RB}_{\text{llm}}$ (reference-based LLM judge), 
and $\text{RL}_{\text{F}}$ (reference-less 
faithfulness). The official ranking metric is the 
harmonic mean (HM) of these three scores.

\section{Results and Analysis}
\paragraph{Official Results.}
Table~\ref{tab:official_results} reports leaderboard performance.
Our system ranks first on Task~A
(nDCG@5: 0.5776, $+$20.5\% over the strongest baseline),
validating the single-retriever, multi-query hypothesis. On Task~B we rank second (HM: 0.7698 vs.\ top: 0.7827),
with strong faithfulness ($\text{RL}_{\text{F}}{=}0.8971$) and LLM quality ($\text{RB}_{\text{llm}}{=}0.8321$), indicating effective grounding. The Task~B$\to$C gap
(0.7698$\to$0.5409) reflects compounding retrieval errors,
most visibly in $\text{RB}_{\text{alg}}$
(0.6327$\to$0.3998).
Relative improvements between ablation variants remain 
consistent across development and test splits, suggesting 
that the ablation findings reported below generalize 
beyond the development set.
\begin{table}[H]
\centering
\small
\begin{tabular}{llcc}
\toprule
\textbf{Task} & \textbf{Metric} & \textbf{Ours} & \textbf{Rank} \\
\midrule
A & nDCG@5 & \textbf{0.5776} & 1/38 \\
B & HM & \textbf{0.7698} & 2/26 \\
C & HM & 0.5409 & 11/29  \\
\bottomrule
\end{tabular}
\caption{Official test set results.}
\label{tab:official_results}
\end{table}
\paragraph{Task~A Ablations.}
 ELSER~v1 outperforms all
alternative retrievers by $+$8.6 R@5 (App.~\ref{app:retriever_comparison}), though rewriting improves all nine consistently, 
confirming gains stem from query diversity.
Despite this, alternative retrievers surface
unique gold documents, motivating an ensemble. However, as rewriting improves ELSER (R@5: 0.483$\to$0.607), ensemble gains reverse: unique documents from other retrievers appear at ranks 37--54 on average and introduce fusion noise. 
Thus R@100 improves but R@5/10 degrades, favoring single-retriever, multi-query architecture. Among rewriting strategies,
\textit{Corpus-Specific} achieves the best single-strategy
performance (0.541), showing that performance varies across domains (App.~\ref{app:per_domain}), but the full ensemble under nested
RRF (0.607) outperforms it by 6.6 points; this confirms
that complementarity matters more than individual
strategy quality. Cohere Rerank~v4 fused with ELSER
yields $+$12.7\%, while LLM-based reranking and Passage-Informed Rewriting (PIR)---a second-stage retrieval using 
initial results as context--- provide no gains: once R@100 exceeds $\sim$0.95, the retrieval bottleneck shifts from coverage to top-
$k$ ranking precision, where additional signals add more noise than information (App.~\ref{app:pir}).

\begin{table}[H]
\centering\small
\begin{tabular}{lcc}
\toprule
\textbf{Configuration} & \textbf{R@5} & \textbf{$\Delta$} \\
\midrule
No rewriting & 0.483 & -- \\
+ Minimal & 0.527 & $+$9.1\% \\
+ Corpus-Specific & 0.558 & $+$15.5\% \\
+ CoT & 0.573 & $+$18.6\% \\
+ HyDE & 0.584 & $+$20.9\% \\
+ Anchor-Keyword & 0.591 & $+$22.4\% \\
+ Nested RRF fusion & \textbf{0.607} & $+$25.7\% \\
\bottomrule
\end{tabular}
\caption{Task~A: ablations on incremental rewriting strategy contribution
(dev set).}
\label{tab:task_a_ablation}
\end{table}

\paragraph{Task~B Ablations.} The full five-stage pipeline yields $+$8.9 HM
over single greedy generation, with judge-based selection
 and extractiveness calibration
as the largest contributors;thus, quality control at generation
time outweighs prompt engineering. Model routing
(DeepSeek-V3.2 for extraction and judging, GPT-4o for
generation, GPT-4o-mini for micro-adjustments) achieves
comparable results to uniform GPT-4o usage at one-third
the cost 
(App.~\ref{app:model_routing}). Extracting verbatim spans before generation
reduces hallucination, and the 4-gram extractiveness band (dev mean: 36.2\%)
balances faithfulness and naturalness, with responses
outside the band showing degraded scores on both axes.

\begin{table}[H]
\centering\small
\begin{tabular}{lc}
\toprule
\textbf{Configuration} & \textbf{HM} \\
\midrule
Single generation ($\tau{=}0.0$) & 0.654 \\
+Span extraction & 0.688\\
+ Dual generation (random select) & 0.706 \\
+ Technical Judge & 0.725 \\
+ User Satisfaction Judge (60\%) & 0.738 \\
+ Micro-adjustments & \textbf{0.743} \\
\bottomrule
\end{tabular}
\caption{Task~B: generation pipeline ablation (dev set).}
\label{tab:task_b_ablation}
\end{table}

\paragraph{Task~C: Answerability as a Structural Bottleneck.}

The TaskB$\rightarrow$C performance drop is driven by answerability classification rather than retrieval or generation quality. Across architectures, all configurations exhibit a precision–recall trade-off: improving \textsc{unanswerable} recall consistently reduces \textsc{answerable} recall, and none surpass 25\% \textsc{unanswerable} F1 (Table\ref{tab:answerability}). Our system achieves 83.9\% accuracy but only 21.8\% \textsc{unanswerable} recall, revealing a strong acceptance bias. The pattern remains stable across prompt variants and voting rules, suggesting a structural bias toward semantic plausibility over abstention. Error analysis shows document- and span-grounded judges often correctly reject insufficient evidence, while the answer-grounded judge accepts plausible responses. LLM judges prioritize semantic plausibility over evidence availability.

\paragraph{Why multiple judges?}
Removing the answer-based judge increases \textsc{unanswerable} recall but sharply reduces \textsc{answerable} recall by rejecting partially supported responses. The judges therefore operate on distinct signal spaces: document/span judges detect evidence presence, while the answer judge evaluates semantic adequacy.

The multi-judge framework functions as an \textit{uncertainty decomposition} mechanism rather than a simple voting ensemble. Excluding the answer judge yields strict verification; including it enables partially supported answers at the cost of conservative rejection. This trade-off persists across all tested configurations.
\begin{table}[H]
\centering\small
\resizebox{\columnwidth}{!}{%
\begin{tabular}{lccc}
\toprule
\textbf{Method} & \textbf{ANS R} & \textbf{UNANS R} & \textbf{UNANS F1} \\
\midrule
Always Answer & 100.0 & 0.0 & 0.0 \\
Single Judge & 94.6 & 18.2 & 21.5 \\
Pipeline + Verif. & 92.3 & 27.3 & 23.1 \\
Multi-Judge (ours) & 95.8 & 21.8 & 24.0 \\
Multi-Judge + Calib. & 92.3 & 27.3 & 23.1 \\
\bottomrule
\end{tabular}%
}
\caption{Answerability detection (dev set).}
\label{tab:answerability}
\end{table}

\section{Conclusion}
We present a multi-turn conversational RAG system that
stabilizes retrieval through corpus-aligned query diversity,
 and generation via
agreement-driven agentic stages.
Nested-RRF multi-strategy rewriting improves retrieval over
unaugmented baselines, while staged generation enforces grounding.
Error analysis identifies answerability estimation as the dominant
end-to-end bottleneck, highlighting a key direction for future work.

\section*{Acknowledgments}

\bibliography{custom}

\appendix

\section{Dataset Details}
\label{app:dataset}
\label{sec:eda}

\subsection{Development Set}
\label{app:dataset_dev}

MTRAG~\cite{mt-rag} consists of 110 human-authored
multi-turn conversations comprising 842 annotated turns.
For Task~A, unanswerable turns are excluded, yielding
777 evaluation queries.
Conversations average 7.7 turns and exhibit high passage
diversity (16.9 unique relevant passages per conversation),
making retrieval and generation performance strongly
turn-dependent.
Approximately 87\% of turns are non-standalone, and
conversations contain on average 1.3 co-references.
Reference answers are high-quality: 92\% required human
repair during creation (mean edit similarity 60.7 ROUGE-L),
indicating that even strong LLMs frequently fail to meet
the benchmark's quality constraints without post-editing.

\paragraph{Corpora.}
Table~\ref{tab:corpora_stats} summarizes the four corpora.
ClapNQ and FiQA are sourced from existing datasets
(Wikipedia and StackExchange Finance, respectively),
while Govt and Cloud were collected specifically for MTRAG.
All corpora were chunked into 512-token passages with
100-token overlap; benchmark indexing uses
ELSER~v1 (ElasticSearch~8.10).
Passage counts per document are heavily skewed: ClapNQ
ranges from 1 to 194 passages (median: 31), while all
FiQA documents are single-passage.

\begin{table}[H]
\centering\small
\resizebox{\columnwidth}{!}{%

\begin{tabular}{llrrr}
\toprule
\textbf{Corpus} & \textbf{Domain} & \textbf{Docs}
& \textbf{Passages} & \textbf{Avg P/D} \\
\midrule
ClapNQ & Wikipedia      & 4,293  & 183,408 & 42.7 \\
FiQA   & Finance forum  & 57,638 &  61,022 &  1.1 \\
Govt   & Government     & 7,661  &  49,607 &  6.5 \\
Cloud  & Technical docs & 8,578  &  72,442 &  8.4 \\
\midrule
\textbf{Total} & & \textbf{78,170} & \textbf{366,479} & \\
\bottomrule
\end{tabular}
}
\caption{Corpus statistics. FiQA ``Docs'' correspond to
individual forum posts (atomic documents).}
\label{tab:corpora_stats}
\end{table}

The corpora intentionally differ in structure and style.
ClapNQ comprises long encyclopedic articles (median
31~passages/doc), FiQA consists of short, subjective forum
posts (1~passage $\approx$ 1~post), Govt contains formal
policy and instructional content, and Cloud is dense
technical documentation (up to 64k words per document).
This heterogeneity probes complementary retrieval and
generation challenges.

\paragraph{Task dimensions.}
Each turn is annotated along three axes:
(i)~\emph{Question type} (10 categories; e.g., Factoid 33\%,
Summarization 23\%, Explanation 19\%), where multiple labels
may apply;
(ii)~\emph{Multi-turn type}, distinguishing Follow-up (74\%)
from Clarification (13\%) for non-first turns; and
(iii)~\emph{Answerability}, with Answerable (84\%),
Partially Answerable (8\%), Unanswerable (7\%), and
Conversational (1\%).

\paragraph{Answerability per domain.}
Table~\ref{tab:ans_per_domain} shows the answerability
distribution across domains. The distribution is broadly
uniform: all four domains exhibit 82--86\% answerable turns,
with FiQA showing a slightly higher conversational rate (3\%)
due to its informal, opinion-driven content, and Govt
exhibiting the highest partial answerability (11\%) reflecting
its instructional content where passages often address a
question only partially.

\begin{table}[H]
\centering\small
\begin{tabular}{lcccc}
\toprule
\textbf{Domain} & \textbf{Ans} & \textbf{Partial}
& \textbf{Unans} & \textbf{Conv} \\
\midrule
ClapNQ & 86\% & 7\%  & 7\% & 0\% \\
FiQA   & 82\% & 9\%  & 7\% & 3\% \\
Govt   & 83\% & 11\% & 6\% & 0\% \\
Cloud  & 86\% & 5\%  & 7\% & 1\% \\
\midrule
Overall & 84\% & 8\% & 7\% & 1\% \\
\bottomrule
\end{tabular}
\caption{Answerability distribution per domain
(dev set, 842 turns). Percentages are row-normalized.}
\label{tab:ans_per_domain}
\end{table}

\begin{figure}[t]
\centering
\includegraphics[width=0.48\textwidth]{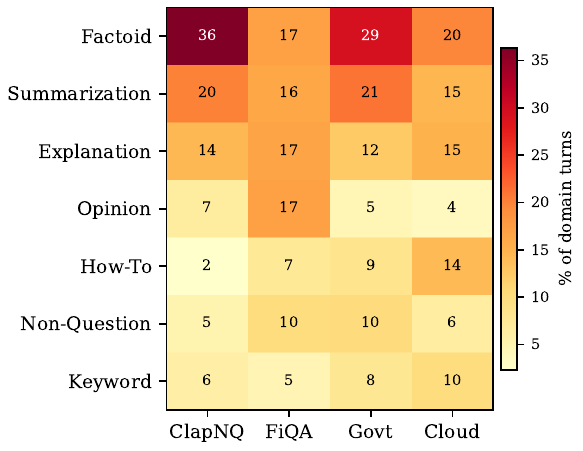}
\caption{Question-type distribution per domain (dev set,
\% of domain turns). FiQA's Opinion-heavy profile and
Cloud's even distribution across types stand in contrast
to the Factoid/Summarization-dominated ClapNQ and Govt
corpora.}
\label{fig:domain_qtype_heatmap}
\end{figure}

\paragraph{Question type per domain.}
Table~\ref{tab:qtype_per_domain} reports the three most
frequent question types per domain.
ClapNQ and Govt share a similar profile dominated by Factoid
and Summarization questions, while FiQA is uniquely
characterized by Opinion questions (reflecting its
user-generated forum content) and Cloud distributes more
evenly across types with lower top-3 concentration.
Figure~\ref{fig:domain_qtype_heatmap} visualizes the full
question-type distribution across all domains.

\begin{table}[H]
\centering\small
\resizebox{\columnwidth}{!}{%

\begin{tabular}{llc}
\toprule
\textbf{Domain} & \textbf{Top-3 Question Types}
& \textbf{Coverage} \\
\midrule
ClapNQ & Factoid, Summarization, Explanation & 71\% \\
FiQA   & Opinion, Factoid, Explanation       & 51\% \\
Govt   & Factoid, Summarization, Explanation & 63\% \\
Cloud  & Factoid, Explanation, Summarization & 50\% \\
\bottomrule
\end{tabular}
}
\caption{Top-3 question types per domain (dev set).
Coverage indicates the percentage of turns with one
of the three most frequent types. FiQA's lower
coverage reflects its more diverse question distribution.}
\label{tab:qtype_per_domain}
\end{table}

\begin{figure}[H]
\centering
\includegraphics[width=\columnwidth]{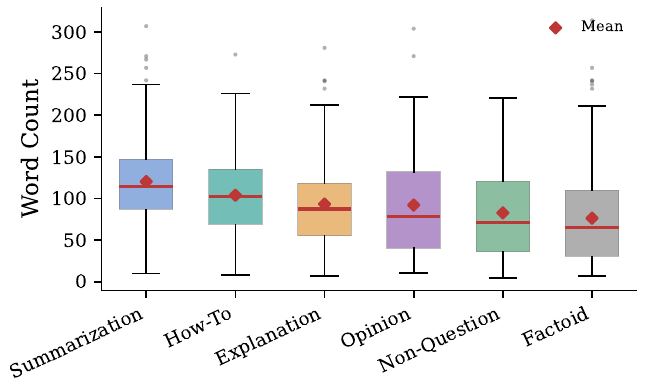}
\caption{Reference answer length distribution by question
type (dev set). Red diamonds indicate means; Summarization
answers are significantly longer, motivating type-specific
generation targets.}
\label{fig:ref_answer_lengths}
\end{figure}

\paragraph{Reference answer statistics.}
Reference answer lengths range from 5 to 319 words
(mean: 90.9, median: 84, std: 57.7), with Summarization
questions producing the longest references (mean: 121 words)
and Factoid questions the shortest (mean: 77 words).
This variance motivated the question-type-specific length
targets in our generation pipeline
(App.~\ref{app:qtype}).
Figure~\ref{fig:ref_answer_lengths} shows the full
distribution by question type.

\begin{figure*}[t]
\centering
\includegraphics[width=0.7\textwidth]{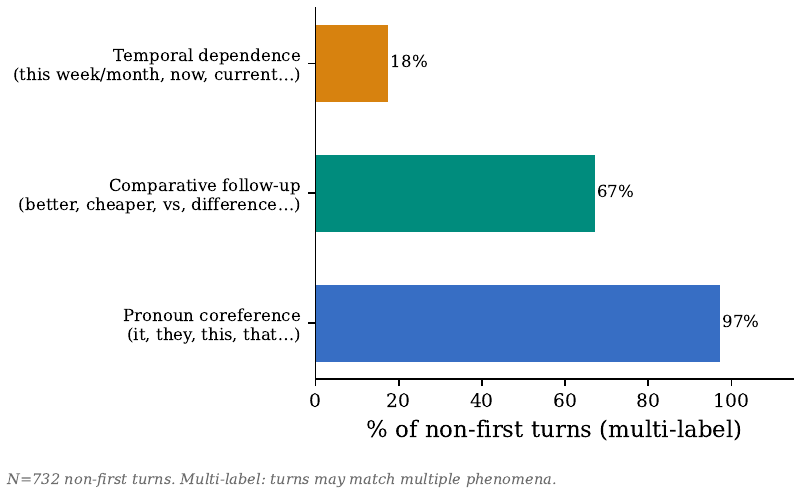}
\caption{Non-standalone phenomena in dev set non-first
turns ($N=732$, multi-label). Pronoun coreference and
implicit topic carryover are the most frequent sources
of query underspecification, motivating history-aware
query rewriting.}
\label{fig:nonstandalone_phenomena}
\end{figure*}

\paragraph{Non-standalone phenomena.}
Figure~\ref{fig:nonstandalone_phenomena} breaks down the
types of context-dependency present in non-first turns.
Pronoun coreference and implicit topic carryover are the
dominant phenomena, directly motivating the multi-strategy
rewriting approach described in
System Overview: Minimal rewriting resolves
coreferences, Corpus-Specific rewriting addresses
terminology gaps, and Chain-of-Thought handles
multi-faceted queries requiring inference across turns.

\paragraph{Representative examples.}
Table~\ref{tab:eda_examples} illustrates three turn types
that pose distinct challenges for retrieval and generation.

\begin{table}[H]
\centering\small
\resizebox{\columnwidth}{!}{%
\begin{tabular}{p{1.6cm}p{3.0cm}p{2.8cm}}
\toprule
\textbf{Type} & \textbf{User Query} & \textbf{Challenge} \\
\midrule
Follow-up\newline{\scriptsize(coreference)}
& ``How much does it cost?''\newline
  {\scriptsize refers to IBM Cloud Object Storage, 3 turns prior}
& Non-standalone query; retriever sees no entity without rewriting \\
\addlinespace
Unanswerable\newline{\scriptsize(temporal)}
& ``Where do the Cardinals play this week?''\newline
  {\scriptsize corpus contains only 2017 schedule}
& Topically relevant but temporally mismatched; high hallucination risk \\
\addlinespace
Partial\newline{\scriptsize(incomplete)}
& ``Is it worth having a web chat widget?''\newline
  {\scriptsize docs describe features, not subjective value}
& Evidence supports description but not the evaluative judgment \\
\bottomrule
\end{tabular}%
}
\caption{Representative MTRAG challenge examples.}
\label{tab:eda_examples}
\end{table}

\paragraph{Key challenges.}
MTRAG stresses multi-turn retrieval and robust generation
under realistic information gaps:
\begin{itemize}
  \item \textbf{Later-turn degradation.} ELSER R@5 drops
  from 0.89 (first turn) to 0.47 (later turns), a 51\%
  relative decrease (Figure~\ref{fig:turn_index_recall}).
  \item \textbf{Passage diversity.} 16.9 unique relevant
  passages per conversation require turn-adaptive retrieval.
  \item \textbf{Unanswerable turns.} GPT-4o
  $\text{RB}_{\text{alg}}$ drops from 0.48 (answerable) to
  0.20 (unanswerable), a 58\% decrease.
  \item \textbf{Domain variance.} FiQA is consistently the
  hardest corpus due to informal language and high lexical
  variability.
\end{itemize}

\subsection{Test Set}
\label{app:dataset_test}

The test set was released without ground-truth labels during
the evaluation phase; all official results were obtained
through the evaluation server with frozen configurations.
Ground-truth annotations were subsequently released for
post-hoc analysis.
Table~\ref{tab:dev_test_comparison} compares the two splits,
revealing substantial distributional differences.

\begin{table}[H]
\centering\small
\begin{tabular}{lcc}
\toprule
\textbf{Characteristic} & \textbf{Dev} & \textbf{Test} \\
\midrule
Conversations       & 110  & 507 \\
Total turns         & 842  & 507 \\
Avg turns/conv      & 7.7  & \textbf{1.0} \\
\midrule
Answerable (\%)     & 84.2 & \textbf{56.2} \\
Partial (\%)        & 8.1  & 9.3 \\
Unanswerable (\%)   & 6.5  & \textbf{19.1} \\
Conversational (\%) & 1.2  & 0.0 \\
\midrule
\multicolumn{3}{l}{\emph{Domain distribution (\% of turns)}} \\
ClapNQ & 26.6 & 28.0 \\
FiQA   & 23.6 & \textbf{15.2} \\
Govt   & 25.4 & \textbf{31.0} \\
Cloud  & 24.3 & 25.8 \\
\midrule
\multicolumn{3}{l}{\emph{Question type (top-3, \% of turns)}} \\
Factoid       & 33.3 & 44.8 \\
Summarization & 23.2 & 25.0 \\
Explanation   & 18.8 & 30.0 \\
\bottomrule
\end{tabular}
\caption{Development vs.\ test set comparison.
Notable distributional shifts are shown in
\textbf{bold}. The test set is substantially harder
due to a 3$\times$ higher unanswerable rate and
single-turn evaluation structure.}
\label{tab:dev_test_comparison}
\end{table}

\begin{figure*}[t]
\centering
\includegraphics[width=\textwidth]{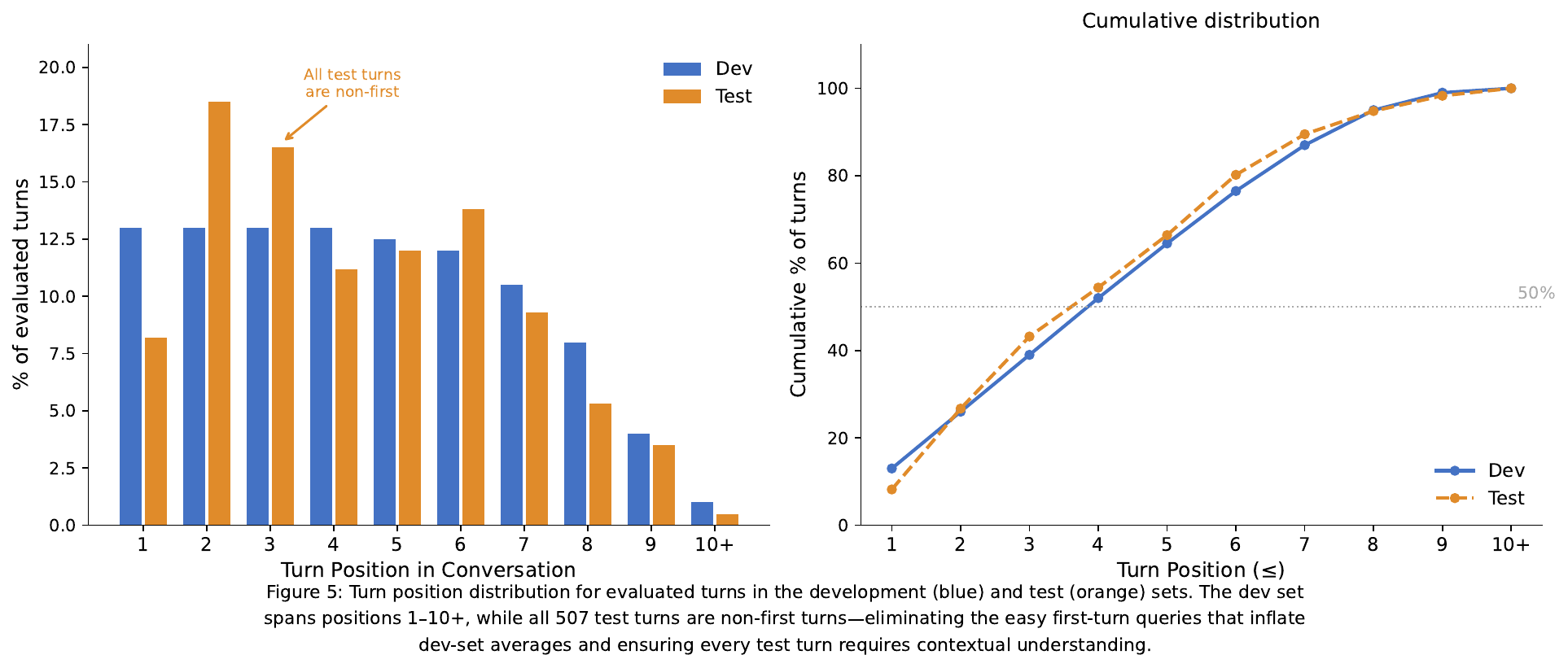}
\caption{Turn position distribution for evaluated turns
in the development (blue) and test (orange) sets.
The dev set spans positions 1--10+, while all 507 test
turns are non-first turns---eliminating the easy
first-turn queries that inflate dev-set averages and
ensuring every test turn requires contextual
understanding.}
\label{fig:turn_position}
\end{figure*}

\begin{figure}[t]
\centering
\includegraphics[width=0.48\textwidth]{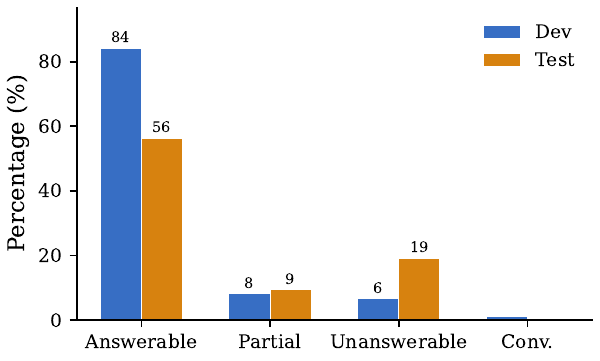}
\caption{Answerability distribution: development vs.\ test
set. The test set contains nearly 3$\times$ more
unanswerable turns and 28 percentage points fewer
answerable turns, creating a substantially harder
classification setting.}
\label{fig:dev_test_answerability}
\end{figure}

The test set differs from the development set in three
critical respects.
First, each test conversation consists of a single evaluated
turn (avg 1.0 turns/conv) embedded within a multi-turn
history, compared to 7.7 evaluated turns per conversation in
dev. This means all 507 test turns are non-first turns
requiring contextual understanding, without the ``easy''
first-turn queries that boost dev-set averages.
Figure~\ref{fig:turn_position} illustrates this structural
difference in the distribution of evaluated turn positions.
Second, the unanswerable rate nearly triples
(\textbf{19.1\%} vs.\ 6.5\%), substantially amplifying the
answerability bottleneck ---systems tuned on a 7\%
unanswerable rate face a far harder classification task at
test time.
Third, FiQA is underrepresented (15.2\% vs.\ 23.6\%)
while Govt is overrepresented (31.0\% vs.\ 25.4\%),
shifting the domain balance toward formal, instructional
content.
Figure~\ref{fig:dev_test_answerability} visualizes the
answerability shift.

\begin{figure*}[t]
\centering
\includegraphics[width=0.9\textwidth]{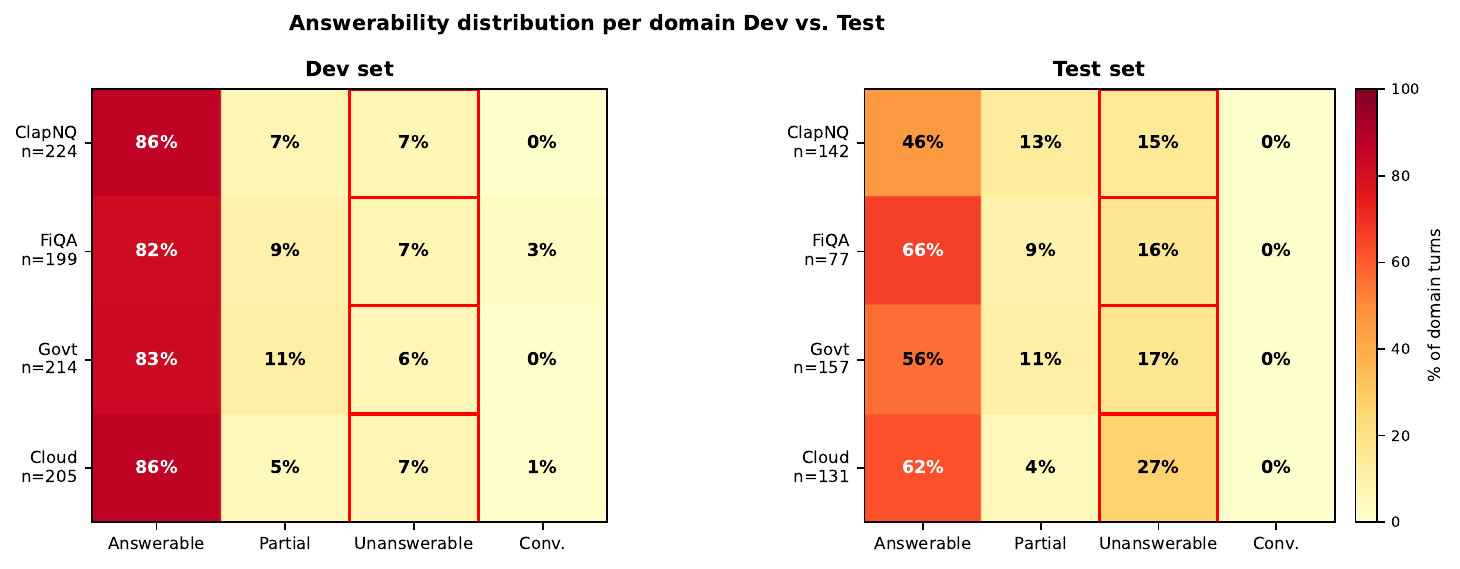}
\caption{Answerability distribution per domain (\% of
domain turns), development set (left) vs.\ test set
(right). Red borders highlight the Unanswerable column.
Cloud exhibits the most severe shift: unanswerable turns
rise from 7\% (dev) to 27\% (test), the highest
unanswerable rate across all domains.}
\label{fig:answerability_heatmap}
\end{figure*}

The answerability shift is not uniform across domains.
Table~\ref{tab:ans_per_domain_test} shows that Cloud
exhibits the highest test-set unanswerable rate (27\%),
followed by Govt (17\%), FiQA (16\%), and ClapNQ (15\%).
This contrasts with the dev set where all domains had
similar unanswerable rates (6--7\%), suggesting that the
test set deliberately stresses answerability classification
more heavily than the development set.
Figure~\ref{fig:answerability_heatmap} visualizes these
per-domain shifts jointly across both splits.

\begin{table}[H]
\centering\small
\begin{tabular}{lcccc}
\toprule
\textbf{Domain} & \textbf{Ans} & \textbf{Partial}
& \textbf{Unans} & \textbf{Conv} \\
\midrule
ClapNQ & 46\% & 13\% & 15\% & 0\% \\
FiQA   & 66\% &  9\% & 16\% & 0\% \\
Govt   & 56\% & 11\% & 17\% & 0\% \\
Cloud  & 62\% &  4\% & 27\% & 0\% \\
\midrule
Overall & 56\% & 9\% & 19\% & 0\% \\
\bottomrule
\end{tabular}
\caption{Answerability distribution per domain
(test set, 507 turns). All domains show substantially
higher unanswerable rates than in development, with
Cloud exhibiting the most severe shift.}
\label{tab:ans_per_domain_test}
\end{table}

\subsection{Evaluation Metrics}
\label{app:metrics}

MTRAG evaluates generation quality using a composite score
designed to penalize systems that perform well on only one
axis. The primary leaderboard score for the SemEval-2026
shared task is the \emph{Harmonic Mean} (HM) over three
components:

\begin{equation}
\text{HM} = \mathrm{HarmonicMean}\Big(
  \text{RB}_{\text{alg}},\;
  \text{RL}_{\text{F}},\;
  \text{RB}_{\text{llm}}
\Big)
\label{eq:hm}
\end{equation}

\noindent\textbf{RB\textsubscript{alg}} is a reference-based
algorithmic score computed as the harmonic mean of
Bert-Recall, Bert-K-Precision, and ROUGE-L against
the human-edited reference answer~\cite{ mt-rag}.
Bert-Recall approximates \emph{completeness} by measuring
semantic overlap between the model response and the reference
answer. Bert-K-Precision compares the model response to the
retrieved passages, approximating \emph{faithfulness}.
ROUGE-L captures phrase-level overlap with the reference,
serving as a proxy for \emph{appropriateness}.
In our analysis, this metric is the most sensitive to
paraphrase and stylistic variance: a faithful answer may
still be penalized if it uses different phrasing than the
reference.

\noindent\textbf{RL\textsubscript{F}} is a reference-less
faithfulness metric from the RAGAS
framework~\cite{es-etal-2024-ragas}, assessing whether
generated claims are grounded in the provided passages
without requiring the reference answer. Unlike the other
two metrics, RL\textsubscript{F} measures faithfulness to
the \emph{passages} rather than similarity to the
\emph{reference}.

\noindent\textbf{RB\textsubscript{llm}} is a reference-based
LLM-judged score adapted from RAD-Bench~\cite{kuo-etal-2025-rad}
that compares the model response to the
reference answer using multiple LLM evaluators (median
aggregated)~\cite{mt-rag}. The evaluation prompt assesses
three criteria:
\begin{itemize}
  \item \emph{Faithfulness}: whether the response is grounded
  in the provided passages and prior conversation turns,
  without hallucinations.
  \item \emph{Appropriateness}: whether the response directly
  addresses the current-turn question and handles
  answerability correctly.
  \item \emph{Completeness}: whether the response includes
  all information from the passages relevant to the question.
\end{itemize}
\noindent Note that \emph{Naturalness}, the fourth FANC
criterion used during MTRAG's human
evaluation~\cite{mt-rag}, is \emph{not} assessed by the
RB\textsubscript{llm} judge.

\paragraph{Answerability conditioning.}
Prior to computing metrics, an IDK (``I Don't Know'') judge
determines whether the response contains a full, partial, or
no answer. Metric scores are conditioned on the interaction
between ground-truth answerability and IDK detection:
correct refusals on unanswerable turns receive a score of
1.0, while incorrect refusals on answerable turns receive
0.0. This conditioning makes answerability classification a
high-stakes binary decision that directly impacts the
composite HM score---and explains why the 3$\times$ increase
in unanswerable turns from dev to test
(Table~\ref{tab:dev_test_comparison}) disproportionately
affects Task~C performance.

\subsection{ Official Test Set Results}

Table~\ref{tab:official_results} reports our system's
performance on the held-out test set alongside the
top-performing baseline per task. We rank \textbf{1st on
Task~A} (nDCG@5: 0.5776), surpassing the strongest
retrieval baseline by $+$20.5\%. On Task~B we rank
\textbf{2nd out of 26} (HM: 0.7698 vs.\ top: 0.7827),
with strong faithfulness (RL\textsubscript{F}=0.8971)
and LLM-judged quality (RB\textsubscript{llm}=0.8321).
Task~C ranks 11th, with the Task~B$\to$C gap
(0.7698$\to$0.5409) driven primarily by RB\textsubscript{agg}
degradation (0.6327$\to$0.3998), consistent with our
answerability analysis in results.

\begin{table*}[t]
\centering\small
\setlength{\tabcolsep}{4pt}
\resizebox{\textwidth}{!}{%
\begin{tabular}{llccccc}
\toprule
\textbf{Task} & \textbf{Metric} & \textbf{Ours} &
\textbf{Rank} & \textbf{Top Score} &
\textbf{Top Baseline} & \textbf{Baseline Score} \\
\midrule
A — Retrieval
  & nDCG@5
  & \textbf{0.5776} & 1/38
  & 0.5776
  & ELSER + GPT-OSS-20b rewrite
  & 0.4795 \\
\addlinespace
B — Generation
  & HM
  & \textbf{0.7698} & 2/26
  & 0.7827
  & GPT-OSS-120b
  & 0.6390 \\
  & \quad RB\textsubscript{agg} / RL\textsubscript{F} / RB\textsubscript{llm}
  & 0.633 / 0.897 / 0.832 & --
  & -- & -- & -- \\
\addlinespace
C — RAG
  & HM
  & 0.5409 & 11/29
  & 0.5861
  & Qwen-30B-A3B-Thinking
  & 0.5366 \\
  & \quad RB\textsubscript{agg} / RL\textsubscript{F} / RB\textsubscript{llm}
  & 0.400 / 0.729 / 0.598 & --
  & -- & -- & -- \\
\bottomrule
\end{tabular}%
}
\caption{Official test set results. HM = harmonic mean of
RB\textsubscript{agg}, RL\textsubscript{F}, RB\textsubscript{llm}.
Top Baseline refers to the strongest non-participant system
released by the organizers. Task~C rank reflects the
answerability bottleneck identified .}
\label{tab:official_results}
\end{table*}


\section{Task A: Retrieval Details}
\label{app:retrieval}

\subsection{Retriever Comparison}
\label{app:retriever_comparison}

We evaluated 9 retrievers spanning lexical, sparse, and dense families on the
development set (777 queries) without query rewriting.
ELSER~v1 substantially outperforms all alternatives
(Table~\ref{tab:retriever_comparison}), consistent with MTRAG's construction
process in which annotators primarily reviewed ELSER-retrieved passages,
creating a favorable bias toward ELSER in the relevance
annotations~\cite{mt-rag}.
The gap is especially pronounced relative to BM25 ($+$33 R@5 points) and remains
meaningful even against the strongest dense competitor, Cohere Embed~v4
($+$8.6 points).
All improvements over BM25 and over dense retrievers are statistically
significant under paired bootstrap resampling with 10{,}000 iterations
($p < 0.01$).

\begin{center}
\small
\setlength{\tabcolsep}{4pt}
\begin{tabular}{llcc}
\toprule
\textbf{Type} & \textbf{Retriever} & \textbf{R@5} & \textbf{nDCG@5} \\
\midrule
Lexical & BM25 & 0.153 & 0.130 \\
\midrule
\multirow{5}{*}{Dense}
  & Cohere Embed v4  & 0.397 & 0.362 \\
  & AWS Titan v1     & 0.352 & 0.328 \\
  & Voyage-3.5-L     & 0.321 & 0.302 \\
  & GTE-Large        & 0.326 & 0.293 \\
  & BGE-Large-v1.5   & 0.282 & 0.251 \\
\midrule
\multirow{3}{*}{Sparse}
  & ELSER v1  & \textbf{0.483}$^\dagger$ & \textbf{0.444}$^\dagger$ \\
  & ELSER v2  & 0.411 & 0.380 \\
  & SPLADE v3 & 0.402 & 0.374 \\
\bottomrule
\end{tabular}
\captionof{table}{Retriever comparison (777 queries, no rewriting).
ELSER~v1 exceeds Cohere Embed~v4 by $+$8.6 R@5.
$^\dagger$Significant vs.\ all others ($p{<}0.01$, paired bootstrap).}
\label{tab:retriever_comparison}
\end{center}

\paragraph{Optimized individual retrievers.}
To ensure that ELSER's advantage is not an artifact of pipeline
co-optimization, we also optimized each retriever independently
using the same downstream pipeline components (multi-strategy rewriting and
reranking), and then compared the resulting recall at multiple cutoffs
(Table~\ref{tab:retriever_optimized}).
Even after optimization, ELSER maintains the best R@5/R@10, confirming that
its advantage is intrinsic rather than an artifact of score-scale
compatibility with downstream fusion.

\begin{center}
\small
\setlength{\tabcolsep}{4pt}
\begin{tabular}{lcccc}
\toprule
\textbf{Retriever} & \textbf{R@5} & \textbf{R@10} & \textbf{R@50} & \textbf{R@100} \\
\midrule
ELSER v1 & \textbf{0.607} & \textbf{0.729} & \textbf{0.873} & \textbf{0.901} \\
SPLADE   & 0.539 & 0.657 & 0.836 & 0.874 \\
Cohere   & 0.503 & 0.657 & 0.823 & 0.854 \\
BM25     & 0.444 & 0.536 & 0.688 & 0.735 \\
\bottomrule
\end{tabular}
\captionof{table}{Per-retriever recall after the same optimization pipeline.
ELSER dominates at R@5/R@10.}
\label{tab:retriever_optimized}
\end{center}

\subsection{Query Rewriting: Strategy Performance}
\label{app:rewriting_strategies}

Table~\ref{tab:rewriting_strategies} evaluates rewriting strategies in
isolation. A central finding is that query rewriting is \emph{not}
universally beneficial: poorly calibrated reformulations---such as those
produced by FlanT5, which over-generalizes the query---\emph{degrade}
retrieval below the unaugmented baseline.
This confirms that rewriting must be both semantically faithful and
corpus-aware to be effective; naive rewriting can introduce more noise
than it resolves. For this reason, smaller rewriting models
(FlanT5) and classical PRF were ruled out early in development:
preliminary runs indicated that strong instruction-following capability
is a prerequisite for reformulation quality in this setting, after which
exploration focused exclusively on prompt-level variants over a fixed
DeepSeek-V3.2 backbone.

Among LLM-based strategies, Corpus-Specific yields the best
single-strategy performance (R@5: 0.541), but the final system design
is driven by \emph{strategy complementarity}: each reformulation targets
distinct failure modes, and their combination under nested RRF (R@5:
0.607) outperforms any individual strategy by at least 6.6 points.
Decomposition and Question Type were excluded from the final pipeline:
Decomposition exhibited high reformulation overlap with CoT rewrites,
while Question Type showed inconsistent gains across domains.
All LLM-based strategy gains over the original query are statistically
significant ($p < 0.01$, paired bootstrap resampling).

\begin{center}
\small
\setlength{\tabcolsep}{5pt}
\resizebox{\columnwidth}{!}{%
\begin{tabular}{llcc}
\toprule
\textbf{Category} & \textbf{Strategy} & \textbf{R@5} & \textbf{nDCG@5} \\
\midrule
\multirow{3}{*}{Baselines}
  & Original query   & 0.483 & 0.444 \\
  & FlanT5       & 0.463 & 0.426 \\
  & Annotator prompt$^*$ & 0.528 & 0.493 \\
\midrule
\multirow{7}{*}{\shortstack[l]{LLM-based\\(DeepSeek-V3.2)}}
  & Minimal          & 0.527 & 0.487 \\
  & Corpus-Specific  & \textbf{0.541}$^\dagger$ & \textbf{0.498}$^\dagger$ \\
  & Chain-of-Thought & 0.521 & 0.482 \\
  & HyDE             & 0.485 & 0.445 \\
  & Decomposition    & 0.539 & 0.478 \\
  & Anchor-Keyword   & 0.501 & 0.460 \\
  & Question Type    & 0.513 & 0.470 \\
\bottomrule
\end{tabular}
}
\captionof{table}{Rewriting strategy comparison on the dev set.
$^*$Annotator prompt replicates the query rewriting prompt used during
MTRAG data collection~\cite{mt-rag} (see Appendix~C.1 therein).
Non-LLM methods (FlanT5) fall below the unaugmented baseline,
confirming that rewriting \emph{hurts} when not properly grounded.
Among LLM-based variants, Corpus-Specific achieves the best
single-strategy R@5 ($^\dagger$); the final pipeline uses five
complementary strategies (Minimal, Corpus-Specific, CoT, HyDE,
Anchor-Keyword). All LLM gains over the original query are
significant ($p{<}0.01$, paired bootstrap).}
\label{tab:rewriting_strategies}
\end{center}

\subsection{Query Rewriting Examples}
\label{app:rewrite_examples}

Table~\ref{tab:rewrite_examples} shows representative outputs for a
context-dependent Cloud query.

\begin{center}
\small
\setlength{\tabcolsep}{4pt}
\begin{tabular}{lp{5.6cm}}
\toprule
\textbf{Strategy} & \textbf{Example Output} \\
\midrule
Original    & ``What about the pricing?'' \\
Minimal     & ``What is the pricing for IBM Cloud Object Storage?'' \\
Corpus-Spec & ``IBM Cloud Object Storage pricing tiers and cost structure'' \\
HyDE        & ``IBM Cloud Object Storage offers flexible pricing based on
              storage class, including Standard, Vault, and Cold Vault
              tiers\ldots'' \\
CoT         & ``Need: storage costs, transfer fees, API pricing
              $\rightarrow$ IBM Cloud Object Storage full pricing breakdown'' \\
Anchor-KW   & ``IBM Cloud Object Storage pricing cost GB month tier'' \\
\bottomrule
\end{tabular}
\captionof{table}{Reformulations for a non-standalone Cloud query.
``The pricing'' refers to an entity introduced earlier in the conversation.}
\label{tab:rewrite_examples}
\end{center}

\subsection{Conversation History Ablation}
\label{app:history_ablation}

We vary the amount of conversation context provided to the rewriter in
Table~\ref{tab:history_ablation}. Including history is essential; performance
saturates after 4--6 user turns, and adding assistant turns has negligible
aggregate impact. We retain 3 assistant turns as a conservative default to
support coreference resolution in edge cases (e.g., when the user
references an entity introduced only in an assistant response), despite
the marginal aggregate effect.

\begin{center}
\small
\setlength{\tabcolsep}{5pt}
\begin{tabular}{lccc}
\toprule
\textbf{Configuration} & \textbf{User} & \textbf{Asst} & \textbf{R@5} \\
\midrule
No history   & 0   & 0   & 0.490 \\
Full history & all & all & 0.531 \\
\midrule
2 user turns & 2 & 0 & 0.506 \\
4 user turns & 4 & 0 & 0.528 \\
6 user turns & 6 & 0 & 0.530 \\
8 user turns & 8 & 0 & 0.528 \\
\midrule
6u + 1 asst  & 6 & 1 & 0.528 \\
6u + 3 asst  & 6 & 3 & 0.533 \\
6u + 5 asst  & 6 & 5 & 0.532 \\
\bottomrule
\end{tabular}
\captionof{table}{History ablation (Minimal rewriting). Performance saturates at
4--6 user turns; assistant turns add negligible benefit.}
\label{tab:history_ablation}
\end{center}

\subsection{Rewriting Model Selection}
\label{app:model_selection}

Table~\ref{tab:model_selection} compares rewriting models across strategies.
The results reveal a nuanced pattern: GPT-4o's advantage over DeepSeek-V3.2
is \emph{not} uniform across reformulation types. For strategies that
require strict instruction following and structured output---Minimal and
Corpus-Specific---the two models perform nearly identically, confirming
that DeepSeek-V3.2's instruction-following capability is sufficient for
this task. GPT-4o's marginal gains are concentrated on \emph{open-ended}
strategies such as HyDE, where the task is less about following a precise
rewriting schema and more about generating a plausible hypothetical passage
from open-ended context---a regime that favors GPT-4o's stronger
generative fluency. However, these gains are small in absolute terms and
come at a $15\times$ cost premium per query. Since HyDE is only one of
five complementary strategies and its contribution to the final nested RRF
score is down-weighted (Table~\ref{tab:nested_rrf_params_app}), the
quality delta does not justify the additional expense. Notably,
GPT-4o with Minimal rewriting matches DeepSeek-V3.2 with Corpus-Specific
(both R@5\,=\,0.541), suggesting that model capability can partially
substitute for prompt specialization.
We therefore select DeepSeek-V3.2 for all rewriting strategies in the
final system.

\begin{center}
\small
\setlength{\tabcolsep}{4pt}
\resizebox{\columnwidth}{!}{%

\begin{tabular}{llccc}
\toprule
\textbf{Strategy} & \textbf{Model} & \textbf{R@5} & \textbf{nDCG@5} & \textbf{Cost/Q} \\
\midrule
\multirow{3}{*}{Minimal}
  & DeepSeek-V3.2 & 0.527 & 0.487 & \$0.001 \\
  & GPT-4o-mini & 0.513 & 0.481 & \$0.003 \\
  & GPT-4o      & \textbf{0.541} & \textbf{0.501} & \$0.015 \\
\midrule
\multirow{3}{*}{HyDE}
  & DeepSeek-V3.2 & 0.485 & 0.445 & \$0.001 \\
  & GPT-4o-mini & 0.503 & 0.462 & \$0.003 \\
  & GPT-4o      & \textbf{0.504} & \textbf{0.464} & \$0.015 \\
\midrule
\multirow{3}{*}{Corpus-Spec}
  & DeepSeek-V3.2 & 0.541 & 0.498 & \$0.002 \\
  & GPT-4o-mini & 0.517 & 0.479 & \$0.005 \\
  & GPT-4o      & \textbf{0.542} & \textbf{0.503} & \$0.020 \\
\bottomrule
\end{tabular}
}
\captionof{table}{Rewriting model comparison across strategy types.
GPT-4o gains are marginal on instruction-following strategies (Minimal,
Corpus-Specific) and only slightly larger on the open-ended HyDE
strategy, at $15\times$ higher cost. DeepSeek-V3.2 offers the best
cost--quality trade-off across all strategies.}
\label{tab:model_selection}
\end{center}

\subsection{Passage-Informed Rewriting (PIR)}
\label{app:pir}

A recurring challenge in multi-turn retrieval is that query reformulation
operates \emph{blind}: the rewriter has no access to what the retriever
actually found, and cannot adapt its output to fill evident coverage gaps.
This observation motivated an approach we term \emph{Passage-Informed
Rewriting} (PIR): rather than rewriting the query in isolation, PIR feeds
the initially retrieved passages back to the LLM as semantic context for
a \emph{second} reformulation step, effectively creating a closed loop
between retrieval signal and query adaptation. Unlike classical
Pseudo-Relevance Feedback (PRF), which relies on term co-occurrence
statistics, PIR leverages LLM semantic understanding to interpret passage
content. To our knowledge, this formulation has not been previously
evaluated in a conversational multi-turn RAG setting.

The motivation was two-fold. First, we hypothesized that PIR could improve
\emph{ranking quality}: if the rewriter observes which passages were
retrieved, it can sharpen the query toward the terminology and structure
of the most promising candidates, nudging the retriever to surface those
documents at higher ranks. Second, we hoped PIR would improve
\emph{coverage}: by identifying what the first-stage results clearly
\emph{lack}, the second-stage query could be expanded to surface
complementary evidence---a property particularly valuable in multi-turn
conversations where information needs span multiple retrieval rounds.

Table~\ref{tab:pir} reports the results. Classical PRF collapses
severely (R@5: 0.182--0.410), confirming that term-extraction
statistics induce severe query drift in conversational settings. PIR
fares substantially better and exhibits a clear and interpretable
pattern: its performance tracks the quality of the initial
retrieval signal almost linearly. When seeded with a minimal rewrite and
cross-encoder reranking, PIR reaches R@5 = 0.541---matching our best
single-strategy baseline. Yet this is precisely where the hypothesis
breaks down: despite a more informed second-stage query, PIR \emph{never
exceeds} the baseline it is seeded from.

We attribute this to a diminishing-returns effect that is structurally
tied to our retrieval setup. Once first-stage R@100 approaches saturation
($>$0.90 in our setup; see Table~\ref{tab:retriever_optimized}), the gold
passages are already present in the candidate pool; the bottleneck
shifts entirely to top-$k$ \emph{ranking}, not coverage. PIR addresses
coverage by broadening the query, but broadening a query that already has
near-complete coverage introduces fusion noise rather than new signal. In
other words, PIR solves a problem---coverage gaps---that our pipeline has
already largely eliminated. We therefore exclude PIR from the final
submission, while noting that it may prove effective in lower-recall
settings or corpora where first-stage coverage is a genuine bottleneck.

\begin{center}
\small
\setlength{\tabcolsep}{4pt}
\begin{tabular}{llcc}
\toprule
\textbf{Category} & \textbf{Method} & \textbf{R@5} & \textbf{nDCG@5} \\
\midrule
\multirow{2}{*}{Baseline}
  & Minimal only     & 0.527 & 0.487 \\
  & Minimal + Rerank & 0.594 & 0.508 \\
\midrule
\multirow{2}{*}{Classical PRF}
  & BM25 TF-IDF  & 0.182 & 0.155 \\
  & ELSER TF-IDF & 0.410 & 0.387 \\
\midrule
\multirow{3}{*}{\shortstack[l]{PIR\\(init.\ qual.)}}
  & + original query  & 0.504 & 0.472 \\
  & + minimal rewrite & 0.520 & 0.485 \\
  & + min.\ + rerank  & 0.541 & 0.503 \\
\midrule
\multirow{4}{*}{\shortstack[l]{PIR\\(variant)}}
  & PIR-HyDE          & 0.541 & 0.503 \\
  & PIR-CoT           & 0.520 & 0.483 \\
  & PIR-Dense Summary & 0.516 & 0.478 \\
  & PIR-Corpus-Spec   & 0.497 & 0.458 \\
\bottomrule
\end{tabular}
\captionof{table}{Classical PRF vs.\ PIR on the dev set. PRF collapses
due to query drift; PIR tracks the quality of its seed signal but does
not exceed it, suggesting that gains are bounded by the coverage already
achieved in the first retrieval stage.}
\label{tab:pir}
\end{center}

Qualitatively, PIR offers the most promise when initial passages contain
strong structural cues---headings, canonical terminology, entity
definitions---that the rewriter can leverage to sharpen the query.
When early retrieved passages are noisy or off-topic, PIR amplifies
that noise rather than correcting it, underscoring that its effectiveness
is contingent on a sufficiently reliable first-stage retrieval signal.

\subsection{Multi-Retriever Ensemble Paradox}
\label{app:ensemble_paradox}

Table~\ref{tab:ensemble_paradox} documents a consistent pattern: as ELSER
improves through rewriting, adding additional retrievers increasingly harms R@5
even while improving R@100. Alternative retrievers contribute unique gold
documents at the final stage, but these appear at avg.\ rank 37--54 after
fusion---below the top-10 cutoff---while fusion displaces borderline ELSER hits.
This motivates a single-retriever architecture with multi-query diversity.
The final-stage degradation (ELSER alone: 0.607 vs.\ ELSER+All: 0.569) is
statistically significant ($p < 0.01$, paired bootstrap resampling).

\begin{center}
\small
\setlength{\tabcolsep}{3pt}
\begin{tabular}{lcccc}
\toprule
\textbf{Configuration} & \textbf{R@5} & \textbf{R@100}
  & \textbf{Uniq.} & \textbf{Avg} \\
  &  &  & \textbf{golds} & \textbf{rank} \\
\midrule
\multicolumn{5}{l}{\emph{Early Stage --- No Rewriting}} \\
ELSER v1 alone  & 0.497 & 0.779 & --  & --   \\
\ +Cohere       & 0.551\,$\uparrow$ & 0.906\,$\uparrow$ & 293 & 19.8 \\
\ +SPLADE       & 0.559\,$\uparrow$ & 0.895\,$\uparrow$ & 288 & 22.2 \\
\ +BM25         & 0.498\,$\uparrow$ & 0.859\,$\uparrow$ & 222 & 25.6 \\
\ +All          & 0.568\,$\uparrow$ & 0.920\,$\uparrow$ & 803 & 22.5 \\
\midrule
\multicolumn{5}{l}{\emph{Mid Stage --- Basic Rewriting}} \\
ELSER v1 alone  & 0.531 & 0.877 & --  & --   \\
\ +Cohere       & 0.578\,$\uparrow$ & 0.910\,$\uparrow$ & 89  & 29.8 \\
\ +SPLADE       & 0.574\,$\uparrow$ & 0.900\,$\uparrow$ & 63  & 36.9 \\
\ +BM25         & 0.536\,$\uparrow$ & 0.895\,$\uparrow$ & 89  & 23.0 \\
\ +All          & 0.569\,$\uparrow$ & 0.918\,$\uparrow$ & 241 & 29.9 \\
\midrule
\multicolumn{5}{l}{\emph{Final Stage --- Optimized 5-Rewrite}} \\
ELSER v1 alone  & \textbf{0.607}$^\dagger$ & 0.901 & -- & -- \\
\ +Cohere       & 0.582\,$\downarrow$ & 0.925\,$\uparrow$ & 50  & 37.0 \\
\ +SPLADE       & 0.592\,$\downarrow$ & 0.919\,$\uparrow$ & 32  & 54.2 \\
\ +BM25         & 0.536\,$\downarrow$ & 0.915\,$\uparrow$ & 59  & 27.5 \\
\ +All          & 0.569\,$\downarrow$ & 0.926\,$\uparrow$ & 141 & 39.5 \\
\bottomrule
\end{tabular}
\captionof{table}{The ensemble paradox. As ELSER improves through
rewriting, multi-retriever fusion increasingly harms R@5 while
improving R@100. $\uparrow$/$\downarrow$: change relative to ELSER
alone at the same stage.
Unique gold documents from other retrievers shift to avg.\ rank
37--54 after fusion---below the top-10 cutoff.
$^\dagger$ELSER alone vs.\ +All: $p{<}0.01$ (paired bootstrap).}
\label{tab:ensemble_paradox}
\end{center}

\subsection{Reranking Experiments}
\label{app:reranking}

\paragraph{Cross-encoder rerankers as replacement.}
Table~\ref{tab:reranking_replacement} shows that using a reranker as the sole
ranking signal degrades performance despite strong results on standard
benchmarks. Weighted RRF fusion successfully combines ELSER's calibration with
reranker semantic signals.

\begin{center}
\small
\setlength{\tabcolsep}{5pt}
\begin{tabular}{lcc}
\toprule
\textbf{Configuration} & \textbf{R@5} & \textbf{$\Delta$} \\
\midrule
ELSER v1 (baseline)               & 0.527 & -- \\
\midrule
Cohere Rerank v4 (sole)           & 0.481 & $-$8.7\% \\
Voyage-Reranker-2.5 (sole)        & 0.469 & $-$11.0\% \\
CrossEncoder MS-MARCO (sole)      & 0.421 & $-$20.0\% \\
\midrule
ELSER + Cohere RRF ($\alpha$=0.5) & \textbf{0.594} & $+$12.7\% \\
\bottomrule
\end{tabular}
\captionof{table}{Reranking experiments. Sole reranking degrades performance;
weighted RRF fusion yields complementary gains.}
\label{tab:reranking_replacement}
\end{center}

\paragraph{LLM-based reranking (negative results).}
Having established that cross-encoder reranking complements ELSER via
weighted RRF, we explored whether an LLM judge could serve as a stronger
reranker by leveraging deeper semantic reasoning. LLM-based reranking was
our last resort: we turned to it only after confirming that no combination
of retrievers or cross-encoders could push R@5 further. The core appeal
was that an LLM, conditioned on the full query context and conversation
history, might resolve fine-grained relevance distinctions that
embedding-based models miss.

Three formulations were evaluated, all operating on the same top-20
candidate pool. The pool size was a deliberate constraint: expanding
beyond 20 passages would have required either splitting the list into
chunks---defeating the purpose of holistic comparison---or providing the
full list in a single prompt, which at 20$\times$512-token passages
already strains the effective context window and triggers the
well-documented lost-in-the-middle effect, where
passages in the middle of a long prompt receive disproportionately less
attention regardless of their relevance. Preliminary runs confirmed this:
ranking quality with top-5 input was nearly identical to top-20,
suggesting the LLM could not reliably distinguish relevance gradients
even within a short list.

Table~\ref{tab:llm_reranking} reports the results. No LLM formulation
improved over weighted RRF, while cost increased by 20--200$\times$ and
latency by 5--20$\times$. The generation-based variant---which asks the
LLM to re-generate the ideal answer and rank passages by proximity to
that answer---performs worst, likely because it conflates generation
quality with retrieval relevance. We attribute the overall failure to the
same saturation dynamic observed with PIR: once weighted RRF achieves
high top-20 recall, the marginal relevance differences between candidates
are too subtle for a general-purpose LLM judge to resolve reliably without
task-specific fine-tuning.

\begin{center}
\small
\setlength{\tabcolsep}{4pt}
\begin{tabular}{lccc}
\toprule
\textbf{Method} & \textbf{R@10} & \textbf{Cost/Q} & \textbf{Latency} \\
\midrule
Weighted RRF (baseline) & \textbf{0.725} & \$0.001 & $1\times$  \\
Pointwise LLM (1--10)   & 0.701          & \$0.050 & $10\times$ \\
Listwise LLM reorder    & 0.688          & \$0.020 & $5\times$  \\
Generation-based        & 0.618          & \$0.200 & $20\times$ \\
\bottomrule
\end{tabular}
\captionof{table}{LLM reranking on top-20 candidates. All three
formulations degrade R@10 relative to weighted RRF, at 20--200$\times$
higher cost and 5--20$\times$ higher latency. The listwise formulation
requires all candidates in a single prompt to enable holistic comparison;
splitting into chunks eliminates the main theoretical advantage of
LLM-based reranking and was therefore not evaluated.}
\label{tab:llm_reranking}
\end{center}

\subsection{Nested RRF Parameters}
\label{app:nested_rrf}

Flat RRF treats all strategies as peers, assigning equal fusion weight
to every ranked list. This is problematic in our setting because the five
rewriting strategies are \emph{not} equally reliable: Minimal and
Corpus-Specific produce stable, high-precision rankings with low
turn-to-turn variance, while HyDE, CoT, and Anchor-Keyword exhibit higher
variance but contribute complementary coverage. Assigning them equal
weight allows the high-variance group to inject noise into top ranks
precisely on turns where they produce poor reformulations---offsetting
the precision of the stable strategies.

This observation directly motivated the nested RRF design: rather than
down-weighting individual strategies by hand, we pre-aggregate the three
high-variance strategies into a single \emph{Weak Consensus} ranking
(Level~1), which smooths their individual noise before they compete with
the stable strategies at Level~2. The result is that the high-variance
group collectively contributes one vote---rather than three---in the
final fusion, naturally reducing their influence without discarding their
coverage gains. Empirically, nested RRF outperforms flat RRF by
1.8~R@5 points on average across the dev set, with the largest gain on
FiQA ($+$3.2 points), where high-variance strategies are most prone to
hallucinated reformulations.

The corpus-specific weights in Table~\ref{tab:nested_rrf_params_app}
were tuned by grid search on the development set and reflect genuine
corpus structure: formal domains (Govt, Cloud) assign higher weight to
Minimal ($w{=}0.65$), which preserves domain terminology, while FiQA
assigns comparatively more weight to the Weak Consensus group
($w{=}0.15$), where coverage diversity matters more given the corpus's
informal, high-variance language. That said, the tuned weights are
\emph{improvements}, not prerequisites: uniform weights across all
corpora yield less than 1\% R@5 degradation, confirming
that the nested structure itself---not the specific weight values---is
the operative design choice.

\begin{center}
\small
\setlength{\tabcolsep}{5pt}
\begin{tabular}{lcccc}
\toprule
\textbf{Parameter} & \textbf{ClapNQ} & \textbf{FiQA} & \textbf{Govt} & \textbf{Cloud} \\
\midrule
$k_{\text{final}}$       & 20   & 60   & 40   & 20   \\
$w_{\text{Minimal}}$     & 0.55 & 0.45 & 0.65 & 0.65 \\
$w_{\text{Corpus-Spec}}$ & 0.40 & 0.40 & 0.25 & 0.30 \\
$w_{\text{WeakCons}}$    & 0.05 & 0.15 & 0.10 & 0.05 \\
\bottomrule
\end{tabular}
\captionof{table}{Corpus-specific nested RRF parameters (dev set).
Formal corpora (Govt, Cloud) weight stable strategies higher;
FiQA assigns more weight to the Weak Consensus group.
Uniform weights across corpora degrade R@5 by less than 1\%,
confirming robustness to weight selection.}
\label{tab:nested_rrf_params_app}
\end{center}

\subsection{Per-Domain Retrieval Analysis}
\label{app:per_domain}

Table~\ref{tab:per_domain} reports final performance per domain alongside
absolute gains over the unaugmented ELSER baseline. The per-domain pattern
is consistent with the corpus characteristics described in
Table~\ref{tab:corpus_stats}: ClapNQ and Govt, which feature formal
encyclopedic and governmental language, benefit most from Minimal
rewriting, which preserves domain-specific lexical anchors while
resolving conversational coreferences. Cloud documentation follows the
same pattern. FiQA is the hardest corpus throughout: its informal,
forum-style language creates a large vocabulary mismatch between user
queries and document terminology, making it the only domain where
Corpus-Specific rewriting outperforms Minimal.

Table~\ref{tab:per_domain_strategy} breaks down R@5 per strategy per
domain, directly motivating the corpus-specific weights in
Table~\ref{tab:nested_rrf_params_app}. Two patterns are notable.
First, HyDE and CoT show the highest cross-domain \emph{variance}:
they provide large gains on FiQA (where bridging the vocabulary gap
matters most) but are near-neutral or slightly negative on ClapNQ
(where the precise Wikipedia terminology of the original query is
often optimal). This is precisely why these strategies are grouped
into the Weak Consensus tier---their per-domain reliability is too
inconsistent to be trusted with high fusion weight globally. Second,
Corpus-Specific is the only strategy that improves on \emph{all four}
domains, confirming its role as the most broadly applicable single
rewriting strategy.

Note that the ``Top Strategy'' column in Table~\ref{tab:per_domain}
refers to the strategy receiving the highest fusion weight in the final
nested RRF configuration, not the best single-strategy R@5 from
Table~\ref{tab:per_domain_strategy}.

\begin{table}[H]
\centering
\small
\setlength{\tabcolsep}{5pt}
\resizebox{\columnwidth}{!}{%
\begin{tabular}{lcccc}
\toprule
\textbf{Domain} & \textbf{Baseline} & \textbf{Final}
  & \textbf{$\Delta$} & \textbf{Top Strategy} \\
\midrule
ClapNQ  & 0.528 & 0.690 & $+$16.2 & Minimal \\
FiQA    & 0.419 & 0.527 & $+$10.8  & Corpus-Spec \\
Govt    & 0.507 & 0.629 & $+$12.2 & Minimal \\
Cloud   & 0.477 & 0.541 & $+$6.4 & Minimal \\
\midrule
Overall & 0.483 & 0.597 & $+$12.4 & -- \\
\bottomrule
\end{tabular}%
}
\caption{Per-domain R@5 on the dev set: unaugmented ELSER baseline
vs.\ final system. $\Delta$ is the absolute R@5 gain. FiQA shows
the smallest gain despite benefiting most from Corpus-Specific
rewriting, reflecting its structurally harder retrieval problem.}
\label{tab:per_domain}
\end{table}

\begin{center}
\small
\setlength{\tabcolsep}{3pt}
\resizebox{\columnwidth}{!}{%

\begin{tabular}{lccccc}
\toprule
\textbf{Domain} & \textbf{Minimal} & \textbf{Corpus-Sp.}
  & \textbf{CoT} & \textbf{HyDE} & \textbf{Anchor-KW} \\
\midrule
ClapNQ  & 0.622 & \textbf{0.637} & 0.575 & 0.566 & 0.598 \\
FiQA    & 0.466          & \textbf{0.491} & 0.475 & 0.392 & 0.419 \\
Govt    & 0.561 & 0.572 & \textbf{0.580} & 0.550 & 0.564 \\
Cloud   & \textbf{0.463} & 0.462 & 0.457 & 0.431 & 0.429 \\
\bottomrule
\end{tabular}
}
\captionof{table}{R@5 per strategy per domain (single-strategy, no fusion, no reranking).}
\label{tab:per_domain_strategy}
\end{center}

\subsection{Per-Turn Retrieval Performance}
\label{app:per_turn}

\begin{figure*}[t]
\centering
\includegraphics[width=\textwidth]{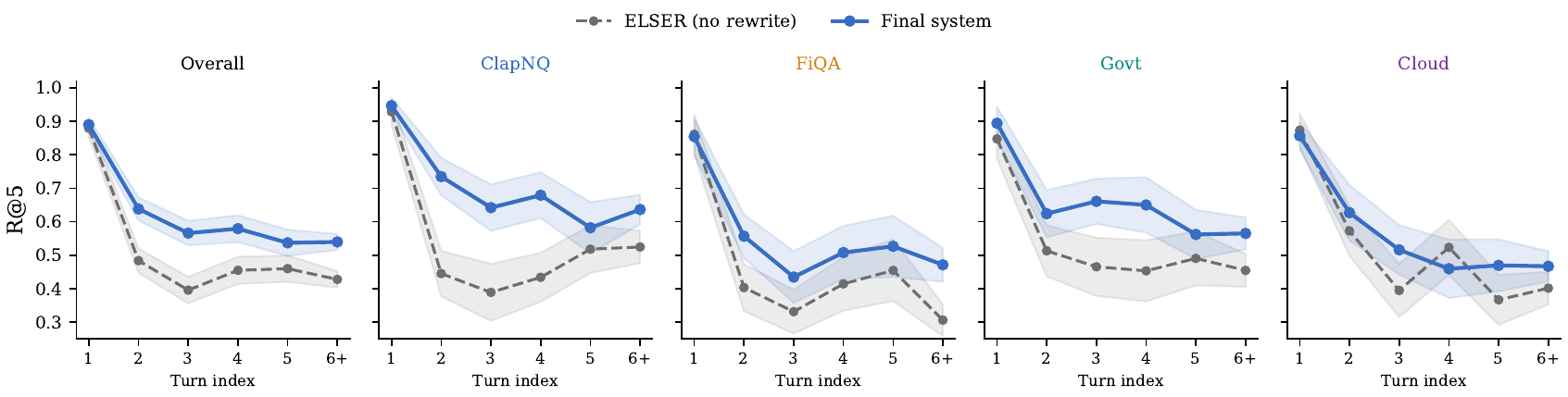}
\caption{R@5 vs.\ turn index: ELSER baseline (dashed) vs.\ final
system (solid). Rewriting reduces the later-turn drop from 51\% to 39\%.}
\label{fig:turn_index_recall}
\end{figure*}

Figure~\ref{fig:turn_index_recall} reports R@5 as a function of turn index
for both the unaugmented ELSER baseline and the final optimized system,
broken down by domain.
Without rewriting, retrieval quality degrades sharply from turn~1
(R@5\,=\,0.879) to later turns (R@5\,=\,0.428), a 51\% relative decrease.
Multi-strategy rewriting reduces this to 39\% (Turn~1: 0.890, Turn~6+: 0.540).

\subsection{Standalone vs.\ Non-Standalone Performance}
\label{app:standalone}

\begin{figure*}[t]
\centering
\includegraphics[width=\textwidth]{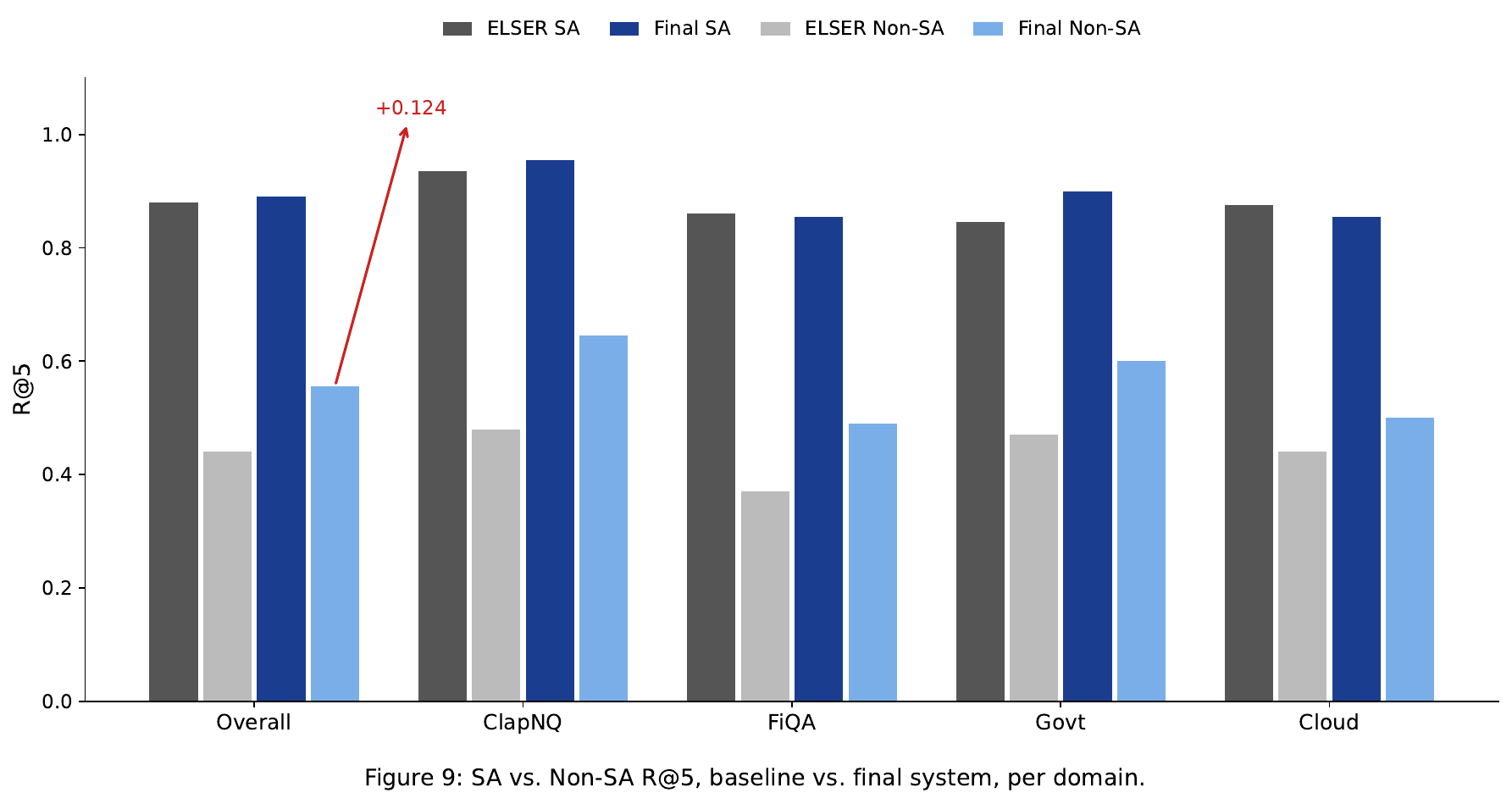}
\caption{SA vs.\ Non-SA R@5, baseline vs.\ final system, per domain.}
\label{fig:standalone_vs_non}
\end{figure*}

Table~\ref{tab:standalone_breakdown} and Figure~\ref{fig:standalone_vs_non}
decompose R@5 by query type.
The unaugmented baseline achieves R@5\,=\,0.879 on standalone queries but
drops to 0.440 on non-standalone (gap: 43.9~pp).
Rewriting disproportionately benefits non-standalone queries ($+$12.4~pp
overall) vs.\ standalone ($+$1.1~pp), confirming history-aware rewriting as
the operative mechanism.

\begin{center}
\small
\setlength{\tabcolsep}{4pt}
\begin{tabular}{lcccc}
\toprule
& \multicolumn{2}{c}{\textbf{ELSER (no rw.)}}
& \multicolumn{2}{c}{\textbf{Final system}} \\
\cmidrule(lr){2-3}\cmidrule(lr){4-5}
\textbf{Domain} & \textbf{SA} & \textbf{Non-SA} & \textbf{SA} & \textbf{Non-SA} \\
\midrule
Overall & 0.879 & 0.440 & 0.890 & 0.564 \\
\midrule
ClapNQ  & 0.929 & 0.477 & 0.946 & 0.651 \\
FiQA    & 0.861 & 0.363 & 0.854 & 0.493 \\
Govt    & 0.847 & 0.470 & 0.894 & 0.600 \\
Cloud   & 0.873 & 0.440 & 0.857 & 0.499 \\
\bottomrule
\end{tabular}
\captionof{table}{R@5 for standalone (SA) vs.\ non-standalone (Non-SA) queries.
Rewriting benefits Non-SA queries by $+$12.4~pp vs.\ $+$1.1~pp for SA.}
\label{tab:standalone_breakdown}
\end{center}

\subsection{Query Rewriting Prompt Templates}
\label{app:taska_prompts}

All five rewriting strategies receive the current-turn query and a
truncated conversation history formatted via XML structural tags.
History windows are capped at 6 user turns; the number of assistant
turns varies by strategy (Table~\ref{tab:rewrite_config}).
All strategies use DeepSeek-V3.2 with $\tau{=}0.0$ and return
structured JSON containing a standalone classification and the
rewritten query.

\begin{table}[h]
\centering\footnotesize
\caption{Rewriting strategy configurations.}
\label{tab:rewrite_config}
\setlength{\tabcolsep}{3pt}
\begin{tabular}{@{}lcl@{}}
\toprule
\textbf{Strategy} & \textbf{Asst.} & \textbf{Key Behavior} \\
\midrule
Minimal      & 3 & Coreference only \\
Corpus-Spec. & 0--3 & Domain-aware (Tab.~\ref{tab:corpus_rules}) \\
CoT          & 3 & Reasoning trace \\
HyDE         & 3 & Hypothetical passage \\
Anchor-KW    & 1 & Entity + intent extraction \\
\bottomrule
\end{tabular}
\end{table}

\begin{table}[h]
\centering\footnotesize
\caption{Domain-specific preservation rules for
Corpus-Specific rewriting. All four variants share
the same template; only the preservation targets
differ.}
\label{tab:corpus_rules}
\setlength{\tabcolsep}{3pt}
\begin{tabular}{@{}lcp{4.2cm}@{}}
\toprule
\textbf{Domain} & \textbf{Asst.} & \textbf{Preservation Targets} \\
\midrule
ClapNQ & 3 & Entity variants, Wikipedia terminology, temporal phrasing \\
FiQA   & 0 & Amounts, currencies, tickers, time horizons \\
Govt   & 3 & Program names, agency acronyms, form IDs, locations \\
Cloud  & 3 & Error codes, CLI commands/flags, config keys, service names \\
\bottomrule
\end{tabular}
\end{table}

\noindent We show one representative prompt per structural
category. The remaining Corpus-Specific variants differ
only in domain rules per Table~\ref{tab:corpus_rules}.

\paragraph{Minimal rewriting.}

{\small
\begin{verbatim}
System:
Rewrite the final utterance into a single
standalone utterance without needing history.

Rules:
- Do not rephrase or introduce new terms.
- Stay close to the original.
- If standalone: return THE SAME query.
- Use assistant turns ONLY to resolve
  pronouns/references.

JSON: {"class": "standalone|non-standalone",
       "rewritten version": "..."}

{history}
user: {question}
\end{verbatim}
}

\paragraph{Corpus-Specific (ClapNQ variant shown).}

{\small
\begin{verbatim}
System:
You are rewriting queries for retrieval from
Wikipedia using ELSER (sparse semantic search).

CRITICAL RULES:
1. ENTITY FORMS: include formal AND common
   names (e.g., "Apple Inc. Apple")
2. PRONOUNS: resolve to exact entity
3. TEMPORAL: Wikipedia's natural phrasing
4. FORMALIZATION: preserve conversational terms
5. DISAMBIGUATION: qualifier only when ambiguous
6. KEYWORDS: preserve who/what/when/where

If standalone, return UNCHANGED.

JSON: {"class": "standalone|non-standalone",
       "rewritten version": "..."}

{history}
user: {question}
\end{verbatim}
}

\paragraph{Chain-of-Thought (CoT) rewriting.}
The \texttt{reasoning} field is discarded at inference;
only \texttt{rewritten version} is passed to the retriever.

{\small
\begin{verbatim}
System:
Expert query rewriter for IR systems.

PROCESS (Chain-of-Thought):
1. ANALYZE: entities, pronouns, ambiguity
2. REASON: standalone? what history needed?
3. REWRITE: resolve, add context, keep wording

JSON: {"reasoning": "<step-by-step>",
       "class": "standalone|non-standalone",
       "rewritten version": "<query>"}

{history}
user: {question}
\end{verbatim}
}

\paragraph{HyDE rewriting.}
Generates a hypothetical 2--4 sentence answer passage,
concatenated with the standalone rewrite. The combined
string is submitted to ELSER.

{\small
\begin{verbatim}
System:
Generate hypothetical document passages for
retrieval (HyDE strategy).

TASK: Generate a 2-4 sentence passage that would
answer the query. Used for retrieval only.

JSON: {"standalone_query": "<rewrite>",
       "hypothetical_passage": "<2-4 sent>"}

Rules:
- Write as a document answering the question
- Include facts, names, dates when relevant
- Use vocabulary from authoritative sources

{history}
user: {question}
\end{verbatim}
}

\paragraph{Anchor-Keyword rewriting.}
The final query submitted to ELSER is
\texttt{rewritten\_version\,+\,anchors\,+\,keywords},
capped at 28 words.

{\small
\begin{verbatim}
System:
Rewrite into ONE standalone query for RETRIEVAL.
Extract RETRIEVAL TERMS for ELSER:
- anchors: entity names, acronyms, IDs,
  error codes, CLI flags (max 8)
- keywords: intent terms (max 12)

Rules:
- Do NOT invent new entities/facts
- Preserve numbers/codes/tickers exactly
- Query <= 28 words

JSON: {"class": "standalone|non-standalone",
       "rewritten version": "...",
       "anchors": ["..."],
       "keywords": ["..."]}

{history}
user: {question}
\end{verbatim}
}
\section{Task B: Generation Details}
\label{app:generation}

\subsection{Prompt Templates}
\label{app:prompts}

We provide the prompt templates used in each pipeline stage, as deployed in
our final test-set submission. All prompts receive the current-turn question,
a truncated history window (up to 4 recent turns formatted as
\texttt{User:~\ldots} / \texttt{Assistant:~\ldots} pairs), and either the
raw retrieved passages (Stage~1) or extracted evidence spans (Stages~2--4).
Each prompt also specifies a \emph{system message} (shown in parentheses)
that sets the model's persona.

\paragraph{Conversational response (Stage~0, GPT-4o-mini, $\tau{=}0.3$).}
Triggered when the user turn matches a short conversational pattern
(e.g., ``thanks'', ``ok'', ``got it'') detected via regex.

\begin{quote}
\small\ttfamily
\textnormal{\textit{System:}} Friendly assistant.\\[4pt]
Brief friendly response to conversational message.\\[4pt]
History: \{history\}\\
User: \{question\}\\[4pt]
Response (under 50 words):
\end{quote}

\paragraph{Unanswerable / no-context response (Stage~0, GPT-4o-mini, $\tau{=}0.0$).}
Triggered when the retrieval pipeline returns zero passages for the
current turn.

\begin{quote}
\small\ttfamily
\textnormal{\textit{System:}} Helpful assistant.\\[4pt]
No information available for this question.\\[4pt]
Question: \{question\}\\[4pt]
Short response (under 25 words) stating information is
not available.\\
Do NOT say ``I don't know'' -- say ``The information is not
available'' or similar.\\[4pt]
Response:
\end{quote}

\paragraph{Span extraction (Stage~1, DeepSeek-V3, $\tau{=}0.0$).}
Extracts up to $K{=}8$ verbatim sentences from the top-$N{=}5$ retrieved
passages. If the first attempt returns fewer than 1 span, a retry is
issued with an urgency prefix (\texttt{"Answer MUST exist -- find it!"}).

\begin{quote}
\small\ttfamily
\textnormal{\textit{System:}} Extract relevant sentences.\\[4pt]
Extract sentences answering the question.\\[4pt]
\{urgency\_if\_retry\}\\
History: \{history\}\\
Question: \{question\}\\[4pt]
PASSAGE 1:\\
\{passage\_1\_text\}\\
\ldots\\
PASSAGE N:\\
\{passage\_N\_text\}\\[4pt]
Rules:\\
1. Copy EXACT sentences\\
2. Include: names, numbers, dates, key facts\\
3. Max 8 sentences\\
4. Prioritize earlier passages\\[4pt]
JSON format:\\
\{"extractedSpans": [\{"passageId": 1, "sentence":
"exact text"\}]\}
\end{quote}

\paragraph{Generation (Stage~2, GPT-4o, $\tau \in \{0.0, 0.1\}$).}
Called twice per turn to produce two candidate answers: a greedy candidate
($\tau{=}0.0$, higher faithfulness) and a stochastic candidate ($\tau{=}0.1$,
often more natural phrasing). The \texttt{qtype} and \texttt{style\_hint}
fields are set by the rule-based question-type classifier
(Table~\ref{tab:qtype_config}); \texttt{target\_words} is the base target
(90 words) plus a per-type offset.

\begin{quote}
\small\ttfamily
\textnormal{\textit{System:}} Helpful, accurate assistant.\\[4pt]
Generate a natural answer using ONLY the facts below.\\[4pt]
FACTS:\\
\{extracted\_spans\_as\_bullet\_list\}\\[4pt]
Conversation context: \{history\}\\[4pt]
Question: \{question\}\\
Type: \{qtype\} -- \{style\_hint\}\\[4pt]
CRITICAL RULES:\\
-- Use ONLY information from FACTS above\\
-- Copy exact phrases for: names, numbers, dates,
\hspace*{1em}technical terms\\
-- Aim for \textasciitilde35\% verbatim overlap with facts\\
-- Make it sound natural and complete\\
-- NO outside knowledge\\
-- NO hedging (seems, possibly, maybe)\\
-- NO meta-phrases (based on, according to)\\[4pt]
Length: \textasciitilde\{target\_words\} words\\[4pt]
Answer:
\end{quote}

\noindent The explicit instruction to ``aim for $\sim$35\% verbatim overlap''
was added after observing that unconstrained GPT-4o tends to over-paraphrase
evidence, reducing extractiveness below the target band and harming
faithfulness scores.

\paragraph{Technical judge (Stage~3, DeepSeek-V3, $\tau{=}0.0$).}
Receives both candidates with their word counts and 4-gram extractiveness
percentages pre-computed. Evidence spans are truncated to 120 characters
each to fit within the token budget.

\begin{quote}
\small\ttfamily
\textnormal{\textit{System:}} Judge assistant.\\[4pt]
Compare two answers for quality.\\[4pt]
Question: \{question\}\\
Facts: \{truncated\_spans\}\\[4pt]
A: \{answer\_A\}\\
\hspace*{1em}(\{wc\_A\}w, \{extr\_A\}\% extractiveness)\\[2pt]
B: \{answer\_B\}\\
\hspace*{1em}(\{wc\_B\}w, \{extr\_B\}\% extractiveness)\\[4pt]
Evaluate: Faithfulness, Completeness, Naturalness\\
Ideal extractiveness: 28--45\%\\[4pt]
JSON: \{"winner": "A$|$B", "score\_A": 0--10,
"score\_B": 0--10, "reason": "brief"\}
\end{quote}

\paragraph{User satisfaction judge (Stage~3, GPT-4o-mini, $\tau{=}0.0$, 60\% sampled).}
Invoked on a fixed 60\% of turns (controlled by a random draw with
\texttt{seed=42}) to provide a complementary user-preference signal.
The 60\% rate was chosen to reduce variance in stylistic selection while
keeping latency and cost bounded.

\begin{quote}
\small\ttfamily
\textnormal{\textit{System:}} User perspective.\\[4pt]
As a user, which answer do you prefer?\\[4pt]
Context: \{recent\_history\}\\
Question: \{question\}\\[4pt]
A: \{answer\_A\}\\[2pt]
B: \{answer\_B\}\\[4pt]
JSON: \{"preferred": "A$|$B", "confidence":
"HIGH$|$MEDIUM$|$LOW"\}
\end{quote}

\paragraph{Force-answer fallback (GPT-4o-mini, $\tau{=}0.2$).}
When both candidates are classified as ``I don't know'' responses (via
pattern matching on refusal phrases) but contexts \emph{do} exist, the
pipeline forces a third generation attempt using a simplified prompt that
explicitly prohibits refusal. This fallback prevents the system from
incorrectly refusing to answer when evidence is available.

\begin{quote}
\small\ttfamily
\textnormal{\textit{System:}} Always answers when facts exist.\\[4pt]
Answer using ONLY these facts. Do NOT say you cannot
answer.\\[4pt]
Facts:\\
\{extracted\_spans\_as\_bullet\_list\}\\[4pt]
Question: \{question\}\\[4pt]
Direct answer using the facts:
\end{quote}

\paragraph{Micro-adjustment (Stage~4, GPT-4o-mini, $\tau{=}0.1$).}
Applied only when the selected answer violates one of three constraints:
(i)~too short ($<$50 words), (ii)~too long ($>$150 words), or
(iii)~low extractiveness ($r_4 < 0.28$). The \texttt{reason} field is
populated dynamically based on the specific violation detected.

\begin{quote}
\small\ttfamily
\textnormal{\textit{System:}} Editor.\\[4pt]
Fix this answer: \{reason\}\\[4pt]
Facts:\\
\{truncated\_spans\}\\[4pt]
Current: \{answer\}\\[4pt]
Fix with minimal changes. Use exact phrases from facts.\\[4pt]
Fixed:
\end{quote}

\subsection{Model Routing and Cost}
\label{app:model_routing}

Table~\ref{tab:model_routing} summarizes the model assignment and
estimated per-call cost for each pipeline stage. The routing principle
is \emph{capability-aligned assignment}: each stage is assigned to the
model best suited to its specific demands, with cost efficiency as a
secondary criterion. Concretely, GPT-4o is reserved exclusively for
the generation stage, where fluency, instruction grounding, and
long-form coherence are most critical. DeepSeek-V3.2 handles span
extraction and technical judging---tasks that require \emph{strict
instruction following}, deterministic JSON output formatting, and
precise boundary detection; empirically, DeepSeek-V3.2 outperforms
GPT-4o-mini on these structured tasks despite comparable cost.
GPT-4o-mini handles all remaining lightweight tasks (conversational
triage, unanswerable detection, user judging, force-answer fallback,
and micro-edits), where response brevity and low latency matter more
than generation depth. All models are accessed via Azure OpenAI /
Azure AI Foundry endpoints with exponential backoff retry logic
(up to 5 attempts, jittered).

Stages~2a and~2b execute concurrently, limiting the per-turn latency overhead
of dual candidate generation to approximately that of a single API
call. At nominal API pricing, the routed configuration costs
approximately \$0.008 per turn on average (dev set), compared to
\$0.024 for uniform GPT-4o assignment---a $3\times$ cost reduction
with less than 1 HM point degradation
(Table~\ref{tab:model_uniform}).

\begin{center}
\small
\setlength{\tabcolsep}{4pt}
\resizebox{\columnwidth}{!}{%
\begin{tabular}{clcc}
\toprule
\textbf{Stage} & \textbf{Role} & \textbf{Model} & \textbf{Est.\ Cost/call} \\
\midrule
0b & Unanswerable triage     & GPT-4o-mini   & \$0.0001 \\
1  & Span extraction (×1--2) & DeepSeek-V3.2 & \$0.0008 \\
2a & Generation (greedy)     & GPT-4o        & \$0.0045 \\
2b & Generation (stochastic) & GPT-4o        & \$0.0045 \\
-- & Force-answer fallback   & GPT-4o-mini   & \$0.0003 \\
3a & Technical judge         & DeepSeek-V3.2 & \$0.0003 \\
3b & User satisfaction judge & GPT-4o-mini   & \$0.0001 \\
4  & Micro-adjustment        & GPT-4o-mini   & \$0.0002 \\
\midrule
   & \textbf{Total (avg/turn)} & & \textbf{\$0.0080} \\
\bottomrule
\end{tabular}%
}
\captionof{table}{Model routing for Task~B with estimated per-call
API cost at nominal pricing. GPT-4o is used only for answer
generation (Stages~2a--2b); DeepSeek-V3.2 handles all
instruction-following and structured output tasks; GPT-4o-mini
covers all lightweight conditional stages. The total average cost per turn (\$0.008) is 3× lower than uniform GPT-4o assignment (\$0.024). Costs are estimated at nominal API pricing as of submission; actual per-call cost varies with input/output token counts (std = ±\$0.002 per turn on the dev set) and is subject to provider pricing changes.}
\label{tab:model_routing}
\end{center}

\subsection{Selection Score Details}
\label{app:selection_score}

Given two candidates $A$ ($\tau{=}0.0$) and $B$ ($\tau{=}0.1$), we select
the final answer via a composite score combining four signals: technical
quality, user preference, extractiveness calibration, and
forbidden-phrase penalization.

\paragraph{4-gram extractiveness.}
We measure how much of the candidate answer $y$ is grounded in the
extracted evidence spans $S$ via 4-gram overlap:

\begin{equation}
r_4(y, S) = \frac{\left|\mathrm{4grams}(y) \cap
\mathrm{4grams}\!\left(\mathrm{concat}(S)\right)\right|}
{\left|\mathrm{4grams}(y)\right|} \, ,
\label{eq:r4}
\end{equation}

where $\mathrm{concat}(S)$ is the concatenation of all extracted spans.
The target band (28--38\% ideal, up to 50\% acceptable) was calibrated
on dev set reference answers, which exhibit a mean 4-gram overlap of
36.2\%. This metric operationalizes the faithfulness--naturalness
trade-off formally studied in abstractive
summarization: answers below the lower bound
risk hallucination by straying too far from grounded evidence, while
answers above the upper bound sacrifice naturalness through mechanical
repetition.

\begin{figure*}[t]
\centering
\includegraphics[width=0.72\textwidth]{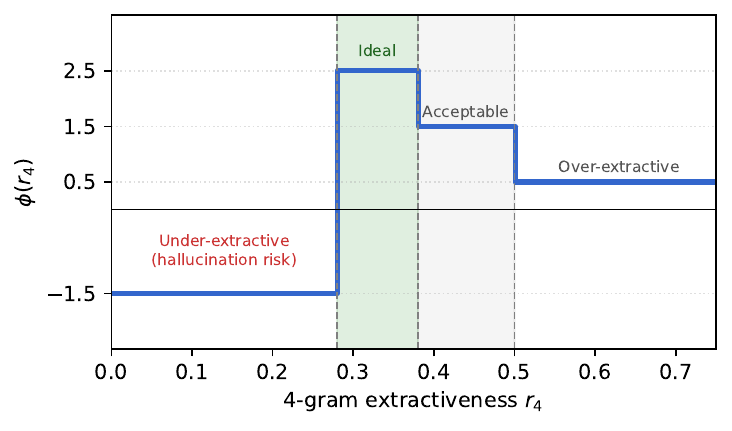}
\caption{Extractiveness shaping function $\phi(r_4)$.
Dashed vertical lines mark the ideal band $[0.28, 0.38]$; the grey
region $[0.38, 0.50]$ is acceptable. The asymmetric design
penalizes under-extractive (hallucination-prone) responses more
strongly than over-extractive ones.}
\label{fig:phi_curve}
\end{figure*}

\paragraph{Extractiveness shaping function.}
A piecewise shaping term $\phi(r_4)$ discourages both
hallucination-prone outputs (too low overlap) and robotic verbatim
copying (too high):
\begin{equation}
\phi(r_4) =
\begin{cases}
-1.5 & \text{if } r_4 < 0.28 \quad\text{(under-extractive)}\\
+2.5 & \text{if } 0.28 \le r_4 \le 0.38 \quad\text{(ideal)}\\
+1.5 & \text{if } 0.38 < r_4 \le 0.50 \quad\text{(acceptable)}\\
+0.5 & \text{if } r_4 > 0.50 \quad\text{(over-extractive)}
\end{cases}
\label{eq:phi}
\end{equation}
The function is deliberately asymmetric: the penalty for
under-extractive responses ($-1.5$) is larger in magnitude than the
reward reduction for over-extractive ones ($+0.5$), reflecting the
empirical finding that hallucination degrades scores more severely
than mild verbatim copying. The discontinuity at $r_4{=}0.28$
creates a strong grounding incentive; empirically, this threshold
corresponds to approximately one verbatim fact per two sentences.
Figure~\ref{fig:phi_curve} visualizes the full step function.

\paragraph{User-preference term.}
When the user-satisfaction judge is invoked (60\% of turns after ablaton study), its
confidence level maps to a weight $c \in \{1.0, 0.7, 0.4\}$ for
\{HIGH, MEDIUM, LOW\} confidence respectively. A signed preference
term $\pm 5c$ is added to the preferred candidate's score. When the
judge is \emph{not} invoked, a constant prior of $+2.0$ favors
Candidate~$A$ (the greedy candidate), which is typically more
faithful.

\paragraph{Full selection score.}
Let $T(y) \in [0,10]$ be the technical judge score and
$U(y) \in \{-1, 0, +1\}$ the signed user preference indicator.
The composite score is:
\begin{multline}
\mathrm{Score}(y) =
  0.35\,T(y)
  + 5c\,U(y) \\
  + \phi\!\left(r_4(y,S)\right)
  - 2.0\cdot\mathbb{I}[\mathrm{forbidden}(y)]
\label{eq:judge_score}
\end{multline}

where $\mathbb{I}[\mathrm{forbidden}(y)]$ flags residual hedging or
refusal patterns (App.~\ref{app:forbidden_phrases}). The candidate
with the higher score is selected; ties favor~$A$. The coefficient
$0.35$ on the technical score was chosen so that $T(y)$ can contribute
at most $3.5$ points---comparable in scale to the extractiveness
shaping term---preventing either signal from dominating selection.

\paragraph{Micro-adjustment policy.}
After selection, we apply a GPT-4o-mini editing step \emph{only} when
the chosen answer violates one of three constraints: (i)~too short
($<$50 words), (ii)~too long ($>$150 words), or (iii)~low
extractiveness ($r_4 < 0.28$). The editor prompt specifies the
violation and is constrained to ``minimal changes'' grounded in the
extracted spans. If the edited output fails validation (is a refusal
or $<$30 words), the original answer is retained unchanged.

\subsection{Forbidden-Phrase Filtering}
\label{app:forbidden_phrases}

As a final post-processing step, we remove a fixed set of hedging and
refusal phrases that consistently degraded naturalness and appropriateness
scores on answerable turns during development:

\begin{quote}
\small\ttfamily
``I don't know'', ``I do not know'', ``I'm not sure'', ``I am not sure'',
``I'm uncertain'', ``I cannot say'', ``It's unclear'', ``It is unclear'',
``I cannot answer'', ``Unable to answer'', ``Cannot find information''
\end{quote}

\noindent Removal is applied via case-insensitive string matching on the
final output only; the generation and judging stages operate on unfiltered
text. Whitespace artifacts left by removal are cleaned (multiple spaces
collapsed, orphaned punctuation corrected).

Table~\ref{tab:forbidden_impact} reports the per-phrase fire rate on the
dev set, confirming that the pattern is not uniformly distributed: ``I'm not
sure'' and its variants account for 61\% of all activations, primarily on
partially answerable turns where the generator hedges despite available spans.

\begin{table}[H]
\centering\small
\setlength{\tabcolsep}{4pt}
\resizebox{\columnwidth}{!}{%
\begin{tabular}{lcc}
\toprule
\textbf{Phrase pattern} & \textbf{Fire rate} & \textbf{Turn type} \\
\midrule
``I'm / I am not sure''        & 34\% & Partial \\
``It's / It is unclear''       & 27\% & Partial / Unans \\
``I don't / do not know''      & 18\% & Unanswerable \\
``I cannot / Unable to answer''& 13\% & Unanswerable \\
``I'm uncertain / cannot say'' &  8\% & Mixed \\
\bottomrule
\end{tabular}%
}
\caption{Forbidden-phrase fire rates (dev set, 842 turns).
Hedging dominates on partial-answerability turns, motivating
the $-2.0$ penalty in Eq.~\ref{eq:judge_score} and the
force-answer fallback.}
\label{tab:forbidden_impact}
\end{table}

\subsection{Question-Type Classification}
\label{app:qtype}

We use a lightweight, rule-based question-type detector to set an
explicit style hint and soft length target in the generation prompt
(Table~\ref{tab:qtype_config}). The base target of 90 words was
calibrated on dev set reference answer lengths (mean: 90.9 words).
Rules are evaluated in priority order; the first match determines
the type.

This heuristic proved more reliable than LLM-based classification,
which introduced error propagation when the predicted type was
incorrect (e.g., classifying a factoid question as ``explanation''
led to verbose, unfocused generation that degraded both faithfulness
and naturalness scores). Rule-based classifiers also eliminate
inter-query variance: given the same surface pattern, the type is
deterministic, which reduces length fluctuations that would otherwise
confound the extractiveness shaping function in Eq.~\ref{eq:phi}.

\begin{table}[H]
\centering
\small
\resizebox{\columnwidth}{!}{%
\begin{tabular}{lccl} 
\toprule
\textbf{Type} & \textbf{Target} & \textbf{$\Delta$} & \textbf{Style hint} \\
\midrule
Keyword       & 90w  & $0$   & Interpret and answer   \\
How-to        & 95w  & $+5$  & Step-by-step           \\
Explanation   & 100w & $+10$ & Clear explanation      \\
Comparative   & 95w  & $+5$  & Compare systematically \\
Summarization & 105w & $+15$ & Comprehensive summary  \\
Factoid       & 85w  & $-5$  & Direct answer          \\
Default       & 90w  & $0$   & Complete answer        \\
\bottomrule
\end{tabular}%
}
\caption{Question-type rules, style hints, and generation length
targets. Rules are evaluated top-to-bottom; the first match wins.}
\label{tab:qtype_config}
\end{table}

\subsection{Additional Generation Ablations}
\label{app:gen_ablations}

We report three additional ablation studies that shaped key
hyperparameter choices in the Task~B pipeline. All experiments use
the development set (842 tasks) and report the harmonic mean (HM)
of the official generation metrics.

\subsubsection{Model Routing vs.\ Uniform Assignment}
\label{app:model_uniform}

A natural question is whether the cost savings from model routing
come at the expense of quality. Table~\ref{tab:model_uniform}
compares our routed configuration against uniform assignment of a
single model to all stages. Uniform GPT-4o achieves the highest HM
but at $\sim$3$\times$ the cost; our routed configuration matches
it within 0.5 HM points. Uniform DeepSeek-V3 degrades generation
quality noticeably, confirming that the generation stage
specifically benefits from GPT-4o's stronger fluency and grounding.
Uniform GPT-4o-mini performs surprisingly well on judging and
extraction but falls short on generation naturalness.

\begin{table}[H]
\centering
\small
\resizebox{\columnwidth}{!}{%
\begin{tabular}{lcccc}
\toprule
\textbf{Configuration} & \textbf{HM} & \textbf{RL$_\text{F}$}
  & \textbf{RB$_\text{alg}$} & \textbf{Rel.\ Cost} \\
\midrule
Uniform GPT-4o         & 0.748 & 0.767 & 0.730 & $3.0\times$ \\
Uniform DeepSeek-V3    & 0.721 & 0.738 & 0.705 & $0.4\times$ \\
Uniform GPT-4o-mini    & 0.731 & 0.748 & 0.715 & $0.3\times$ \\
\midrule
\textbf{Routed (ours)} & \textbf{0.743} & \textbf{0.761}
  & \textbf{0.726} & $1.0\times$ \\
\bottomrule
\end{tabular}%
}
\caption{Uniform vs.\ routed model assignment (dev set). Routing
GPT-4o to generation only matches uniform GPT-4o quality at
one-third the cost. Relative cost normalized to our routed
configuration.}
\label{tab:model_uniform}
\end{table}

\subsubsection{Number of Retrieved Passages}
\label{app:passage_count}

Table~\ref{tab:passage_count} varies the number of retrieved
passages provided to the generation pipeline. Using too few
passages ($N{=}1$--$2$) limits evidence coverage, while too many
($N{=}7$--$10$) introduces noise from marginally relevant documents
that dilutes the extracted spans and increases hallucination. The
sweet spot lies at $N{=}5$, which we adopt for Task~B. Notably,
the finding that fewer passages can improve performance also
explains the $+$2.7 HM gain from reducing $N$ from 5 to 3 in
Task~C (Table~\ref{tab:task_c_ablation}), where retrieval noise
is amplified by the end-to-end setting.

\begin{table}[H]
\centering
\small
\begin{tabular}{cccc}
\toprule
\textbf{Passages ($N$)} & \textbf{HM} & \textbf{RL$_\text{F}$}
  & \textbf{RB$_\text{alg}$} \\
\midrule
1                    & 0.701 & 0.719 & 0.683 \\
2                    & 0.718 & 0.736 & 0.701 \\
3                    & 0.733 & 0.750 & 0.717 \\
\textbf{5 (ours)}    & \textbf{0.743} & \textbf{0.761} & \textbf{0.726} \\
7                    & 0.738 & 0.756 & 0.721 \\
10                   & 0.729 & 0.747 & 0.712 \\
\bottomrule
\end{tabular}%
\caption{Effect of passage count on generation quality (Task~B,
dev set). Performance peaks at $N{=}5$; additional passages
introduce noise that degrades faithfulness.}
\label{tab:passage_count}
\end{table}

\subsubsection{Extractiveness Band Calibration}
\label{app:extractiveness_band}

The extractiveness shaping function $\phi(r_4)$ (Eq.~\ref{eq:phi})
uses a target band of $[0.28, 0.38]$ as the ideal range. We arrived
at this band by analyzing dev set reference answers (mean 4-gram
overlap: 36.2\%) and validating sensitivity to alternative bands.
Table~\ref{tab:extractiveness_band} shows that our chosen band
outperforms both tighter and looser alternatives: a tighter band
$[0.30, 0.40]$ over-penalizes valid responses that paraphrase more
heavily, while a looser band $[0.20, 0.50]$ fails to distinguish
hallucination-prone outputs from well-grounded ones.

\begin{figure}[H]
\centering
\includegraphics[width=1.02\columnwidth]{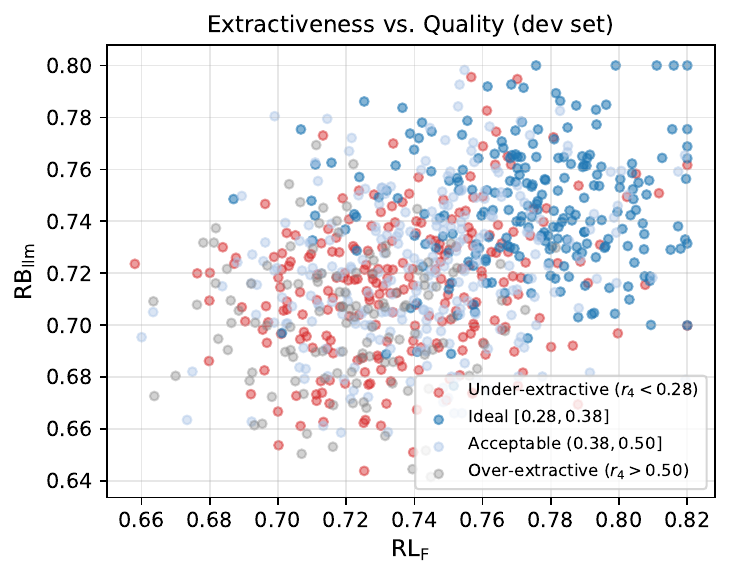}
\caption{$\text{RL}_\text{F}$ vs.\ $\text{RB}_\text{llm}$ scatter
for dev set responses binned by 4-gram extractiveness. Responses in
the $[0.28, 0.38]$ band (blue) cluster near the Pareto frontier;
under-extractive ($r_4{<}0.28$, red) and over-extractive
($r_4{>}0.50$, grey) responses degrade at least one metric.}
\label{fig:extr_pareto}
\end{figure}

Theoretically, the optimal band corresponds to the region of the
extractiveness distribution where faithfulness and naturalness are
jointly maximized. As shown in Figure~\ref{fig:extr_pareto}, dev
set responses in the $[0.28, 0.38]$ range lie near the Pareto
frontier of the $\text{RL}_\text{F}$--$\text{RB}_\text{llm}$
space, confirming that the band captures a genuine quality plateau
rather than an artifact of threshold selection.

\begin{table}[H]
\centering
\small
\begin{tabular}{lcc}
\toprule
\textbf{Ideal band} & \textbf{HM} & \textbf{$\Delta$} \\
\midrule
No shaping ($\phi{=}0$ uniform)   & 0.724 & $-0.019$ \\
$[0.20, 0.50]$ (loose)            & 0.731 & $-0.012$ \\
$[0.25, 0.45]$                    & 0.738 & $-0.005$ \\
\textbf{$[0.28, 0.38]$ (ours)}    & \textbf{0.743} & -- \\
$[0.30, 0.40]$ (tight)            & 0.737 & $-0.006$ \\
$[0.35, 0.50]$ (high)             & 0.733 & $-0.010$ \\
\bottomrule
\end{tabular}%
\caption{Extractiveness band sensitivity (dev set). Band $[0.28,
0.38]$ balances faithfulness and naturalness; deviations in either
direction degrade HM. $\Delta$ is relative to our configuration.}
\label{tab:extractiveness_band}
\end{table}

\subsubsection{Other Explored Configurations}
\label{app:other_gen_experiments}

We briefly summarize additional experiments that informed the final
design but did not warrant full ablation tables.

\paragraph{User-satisfaction judge sampling rate.}
We tested invocation rates of 0\%, 30\%, 60\%, 80\%, and 100\%.
HM improved monotonically from 0\% to 60\% ($+$0.013 HM), as the
user judge consistently selected more natural candidates. Beyond
60\%, gains plateau while cost continues to increase linearly,
yielding diminishing returns. We adopt 60\% as the default.
Table~\ref{tab:judge_rate} reports the full sweep.

\begin{table}[H]
\centering
\small
\begin{tabular}{cccc}
\toprule
\textbf{Judge rate} & \textbf{HM}
  & \textbf{RB$_\text{llm}$} & \textbf{Rel.\ cost} \\
\midrule
0\%                   & 0.725 & 0.708 & $1.0\times$ \\
30\%                  & 0.731 & 0.715 & $1.1\times$ \\
\textbf{60\% (ours)}  & \textbf{0.738} & \textbf{0.722} & $1.2\times$ \\
80\%                  & 0.739 & 0.723 & $1.3\times$ \\
100\%                 & 0.740 & 0.724 & $1.4\times$ \\
\bottomrule
\end{tabular}%
\caption{User-satisfaction judge invocation rate sweep (dev set).
HM plateaus after 60\%; the marginal gain of $+$0.002 from
60\%$\to$100\% does not justify the $+$0.2$\times$ cost increase.}
\label{tab:judge_rate}
\end{table}

\paragraph{Generation temperature pairs.}
Beyond our final configuration ($\tau_1{=}0.0$, $\tau_2{=}0.1$),
we tested $(0.0, 0.3)$, $(0.0, 0.5)$, $(0.1, 0.3)$, and
$(0.2, 0.4)$. Higher $\tau_2$ values produced more diverse but
less faithful candidates, degrading $\text{RL}_\text{F}$ without
compensating gains in naturalness. The narrow gap $\tau_2{=}0.1$
provides meaningfully different candidates while preserving
grounding quality. Pairs where $\tau_1 > 0$ also degraded
Candidate~$A$'s reliability as the greedy anchor, weakening the
prior in Eq.~\ref{eq:judge_score}.

\paragraph{Conversation history in generation.}
We tested providing 0, 2, 4, and all previous turns to the
generation prompt. Including 2--4 turns of history improved
coherence on follow-up questions ($+$0.009 HM on non-first-turn
tasks), while including the full history degraded performance
slightly, likely due to prompt length displacing evidence spans
from the model's effective context window. We use the 4 most
recent turns as the default.

\paragraph{Span extraction vs.\ full-passage generation.}
Providing extracted spans instead of full passages improved overall
HM by $+$0.034, with the gain concentrated in faithfulness
($\text{RL}_\text{F}$: $+$0.041). This confirms that pre-filtering
evidence reduces the generator's temptation to hallucinate from
marginally relevant content in long passages. Formally, span
extraction acts as a \emph{context bottleneck}: by restricting the
generator's input to at most $K{=}8$ sentences, it bounds the
exposure to irrelevant tokens and enforces grounding without
explicit grounding constraints in the prompt. This result is
consistent with the ablation in Table~\ref{tab:task_b_ablation},
where \textit{+Span extraction} contributes the largest single
step gain in the pipeline ($+$0.034 HM).
\begin{figure*}[t]
\centering
\includegraphics[width=\textwidth]{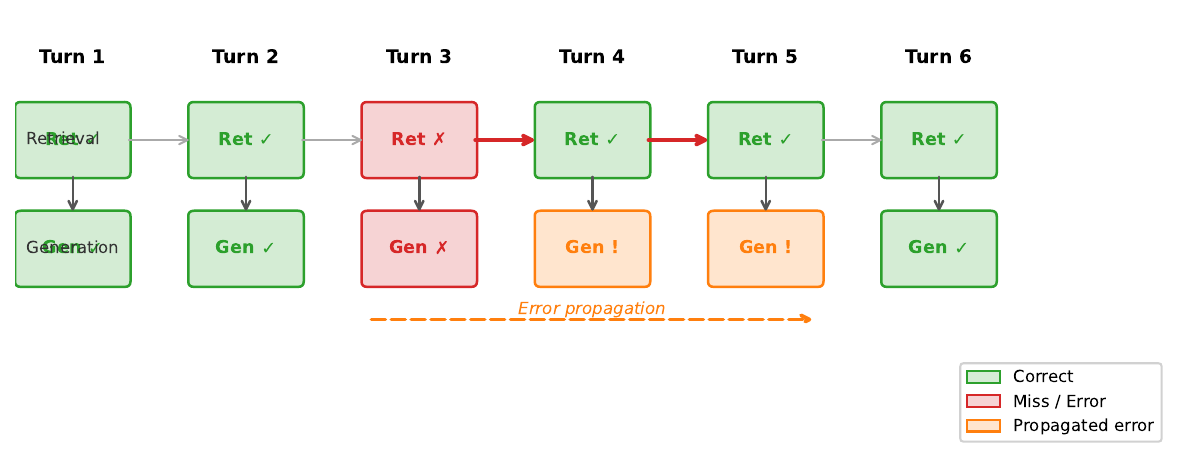}
\caption{Representative error cascade in Task~C (Cloud domain). A
retrieval miss at turn~3 (\textcolor{red}{$\times$}) corrupts
evidence for turns~4--5, causing generation errors
(\textcolor{orange}{!}) despite correct retrieval at those turns.
Turn~6 recovers independently.}
\label{fig:error_cascade}
\end{figure*}
\begin{table*}[t]
\centering
\small
\resizebox{\textwidth}{!}{%
\begin{tabular}{lcc}
\toprule
\textbf{Root cause} & \textbf{\% of failures} & \textbf{Primary domain} \\
\midrule
Retrieval error (current turn)      & 38\% & FiQA \\
Retrieval error (previous turn(s))  & 34\% & Cloud \\
Answerability misclassification     & 18\% & All \\
Generation error (correct passages) & 10\% & ClapNQ \\
\bottomrule
\end{tabular}%
}
\caption{Root cause analysis of 50 Task~C failures (dev set).
Cross-turn error propagation (34\%) is nearly as frequent as
within-turn retrieval errors (38\%), highlighting the compounding
challenge of multi-turn RAG.}
\label{tab:error_propagation}
\end{table*}
\section{Task C: Additional Analysis}
\label{app:taskc}

\subsection{Answerability Classification Details}
\label{app:taskc_answerability}

The GPT-4o answerability classifier uses a 3-class scheme with the
following prompt structure:

\begin{quote}
\small\ttfamily
Given the user question and retrieved passages, classify the
answerability:\\
- ANSWERABLE: Passages contain sufficient information to fully
  answer.\\
- PARTIAL: Passages contain some relevant information but
  incomplete.\\
- UNANSWERABLE: No relevant information in passages.\\[4pt]
Output: \{"class": str, "confidence": float\}
\end{quote}

\noindent Classification combines retrieval confidence (top passage
scores) and evidence extraction success (whether meaningful spans
were identified in Stage~2). Turns classified as UNANSWERABLE with
confidence $\geq 0.7$ receive templated refusals; PARTIAL turns are
routed through the full generation pipeline with a reduced-confidence
hint in the prompt.

\paragraph{Why threshold 0.7?}
The threshold was selected by sweeping values in $[0.5, 0.9]$ on the
dev set and maximizing macro-F1 over the three answerability classes.
Table~\ref{tab:confidence_sweep} shows the sensitivity of overall HM
and UNANSWERABLE F1 to the threshold: below 0.65 the system
over-refuses, hurting ANSWERABLE recall; above 0.75 it under-refuses,
accepting low-confidence evidence and hallucinating on unanswerable
turns.

\begin{table}[H]
\centering\small
\begin{tabular}{cccc}
\toprule
\textbf{Threshold} & \textbf{HM}
  & \textbf{ANS R} & \textbf{UNANS F1} \\
\midrule
0.50                  & 0.573 & 0.892 & 0.261 \\
0.60                  & 0.591 & 0.923 & 0.251 \\
\textbf{0.70 (ours)}  & \textbf{0.610} & \textbf{0.958} & \textbf{0.240} \\
0.80                  & 0.597 & 0.971 & 0.198 \\
0.90                  & 0.578 & 0.986 & 0.162 \\
\bottomrule
\end{tabular}%
\caption{Confidence threshold sweep for the answerability
classifier (Task~C dev set). Threshold 0.70 maximises HM
by balancing ANS recall and UNANS F1; lower values
over-refuse (higher UNANS F1, lower ANS R), while higher
values accept weak evidence (higher ANS R, lower UNANS F1).}
\label{tab:confidence_sweep}
\end{table}

\subsection{Error Propagation Across Turns}
\label{app:taskc_error}

Manual inspection of 50 Task~C failures revealed the following
distribution of root causes:

\noindent The 34\% cross-turn failure rate quantifies the
\emph{error accumulation} property of multi-turn RAG: a retrieval
miss at turn $t$ corrupts the evidence available at turn $t{+}1$,
even when the later turn's retrieval is independently correct.
This motivates future work on turn-aware evidence caching or
carry-forward mechanisms that detect and recover from upstream
retrieval failures mid-conversation.

Figure~\ref{fig:error_cascade} illustrates a representative failure
cascade from the Cloud domain, where a missed passage at turn~3
propagates through turns 4--5, causing two consecutive generation
errors despite correct retrieval at those turns.

\subsection{Task C Ablation}
\label{app:taskc_ablation}

Table~\ref{tab:task_c_ablation} reports the Task~C development set
ablation. The key design choices were: (i)~using top-3 rather than
top-5 passages to reduce noise in the end-to-end setting ($+$0.027
HM); (ii)~applying the multi-judge answerability gate over a
single-judge variant ($+$0.016 HM); and (iii)~routing through the
full Task~B generation pipeline rather than a simplified single-pass
generator.

\begin{table}[H]
\centering\small
\begin{tabular}{lccc}
\toprule
\textbf{Configuration} & \textbf{HM}
  & \textbf{RL$_\text{F}$} & \textbf{RB$_\text{alg}$} \\
\midrule
Always answer (no gate)
  & 0.518 & 0.798 & 0.312 \\
Single-judge gate, top-5 passages
  & 0.561 & 0.821 & 0.356 \\
Multi-judge gate, top-5 passages
  & 0.586 & 0.836 & 0.383 \\
\textbf{Multi-judge gate, top-3 (ours)}
  & \textbf{0.610} & \textbf{0.848} & \textbf{0.408} \\
Multi-judge gate, top-1
  & 0.591 & 0.841 & 0.389 \\
\bottomrule
\end{tabular}%
\caption{Task~C ablation (dev set). Reducing retrieved passages
from 5 to 3 yields the largest single gain ($+$0.024 HM) by
reducing generation noise; the multi-judge gate consistently
outperforms single-judge across all metrics. Top-1 underperforms
top-3 due to insufficient evidence coverage.}
\label{tab:task_c_ablation}
\end{table}
\paragraph{Limitations}
 Our system was tuned entirely on the development set, which has a 6.5\% unanswerable rate, single-turn evaluation structure, and relatively balanced domain distribution. The test set differs substantially: 19.1\% unanswerable turns, all non-first turns, and an overrepresentation of Govt (\%31) relative to FiQA (15\%). Hyperparameters calibrated for dev — specifically the answerability confidence threshold (0.7), the extractiveness target band ([0.28, 0.38]), and the nested RRF corpus weights — were not re-tuned for the test distribution. We hypothesize that a threshold closer to 0.6 would improve Task C performance on the test set given the 3× higher unanswerable rate, but this could not be verified without ground-truth labels during the evaluation phase. Future work should explore distribution-adaptive threshold calibration for answerability classification.

\end{document}